\documentclass[lettersize,journal]{IEEEtran}
\usepackage{amsmath,amsfonts}
\usepackage{algorithmic}
\usepackage{algorithm}
\usepackage{array}
\usepackage[caption=false,font=normalsize,labelfont=sf,textfont=sf]{subfig}
\usepackage{textcomp}
\usepackage{stfloats}
\usepackage{url}
\usepackage{verbatim}
\usepackage{graphicx}
\usepackage{cite}
\usepackage{booktabs} % newly added by author
\usepackage{multirow} % newly added by author
\usepackage{algorithm} % newly added by author
\usepackage{algorithmic} % newly added by author
\usepackage{xcite} % newly added by author
\externalcitedocument{appendix} % newly added by author

\newtheorem{theorem}{Theorem}

\newtheorem{definition}{Definition}

\begin{document}

\title{Dual-Label Learning With Irregularly Present Labels}

\author{Mingqian Li, Qiao Han, Ruifeng Li, Yao Yang,
 Hongyang Chen, ~\IEEEmembership{Senior Member,~IEEE,}
        % <-this % stops a space
\IEEEcompsocitemizethanks{
    % \IEEEcompsocthanksitem This work is supported by XXX (Corresponding author: Hongyang Chen)
    \IEEEcompsocthanksitem  (Corresponding author: Hongyang Chen)
    \IEEEcompsocthanksitem Mingqian Li is with the Research Center for Data Hub and Security, Zhejiang Lab, Hangzhou, China. Email: mingqian.li@zhejianglab.com.
    \IEEEcompsocthanksitem Qiao Han is with the Research Center for Data Hub and Security, Zhejiang Lab, Hangzhou, China. Email: hanq@zhejianglab.com.
    \IEEEcompsocthanksitem Ruifeng Li is with College of Computer Science and Technology, Zhejiang University, Hangzhou, China. Email: lirf@zju.edu.cn.
    \IEEEcompsocthanksitem Yao Yang is with the Research Center for Data Hub and Security, Zhejiang Lab, Hangzhou, China. Email: yangyao@zhejianglab.com.
    \IEEEcompsocthanksitem Hongyang Chen is with the Research Center for Data Hub and Security, Zhejiang Lab, Hangzhou, China. Email: dr.h.chen@ieee.org.}
% <-this % stops a space
\thanks{Manuscript received April 19, 2021; revised August 16, 2021.}
}

\markboth{Journal of \LaTeX\ Class Files,~Vol.~14, No.~8, August~2021}%
{Shell \MakeLowercase{\textit{et al.}}: A Sample Article Using IEEEtran.cls for IEEE Journals}


\maketitle

\begin{abstract}
In multi-task learning, labels are often missing irregularly across samples, which can be fully labeled, partially labeled or unlabeled. The irregular label presence often appears in scientific studies due to experimental limitations. It triggers a demand for a new training and inference mechanism that could accommodate irregularly present labels and maximize their utility. This work focuses on the two-label learning task and proposes a novel training and inference framework, \underline{\textbf{D}}ual-\underline{\textbf{L}}abel \underline{\textbf{L}}earning (\textbf{DLL}). 
    The DLL framework formulates the problem into a dual-function system, in which the two functions should simultaneously satisfy standard supervision, structural duality and probabilistic duality. 
    DLL features a dual-tower model architecture that allows for explicit information exchange between labels, aimed at maximizing the utility of partially available labels. 
    During training, missing labels are imputed as part of the forward propagation process, while during inference, labels are predicted jointly as unknowns of a bivariate system of equations.
    Our theoretical analysis guarantees the feasibility of DLL, and extensive experiments are conducted to verify that by explicitly modeling label correlation and maximizing label utility, our method makes consistently better prediction than baseline approaches by up to 9.6\% gain in F1-score or 10.2\% reduction in MAPE. Remarkably, DLL maintains robust performance at a label missing rate of up to 60\%, achieving even better results than baseline approaches at lower missing rates down to only 10\%.
\end{abstract}

\begin{IEEEkeywords}
Multi-label learning, dual learning, missing labels, multi-task learning.
\end{IEEEkeywords}

\section{Introduction}
\begin{figure}[t]
\centering
\includegraphics[width=0.98\columnwidth]{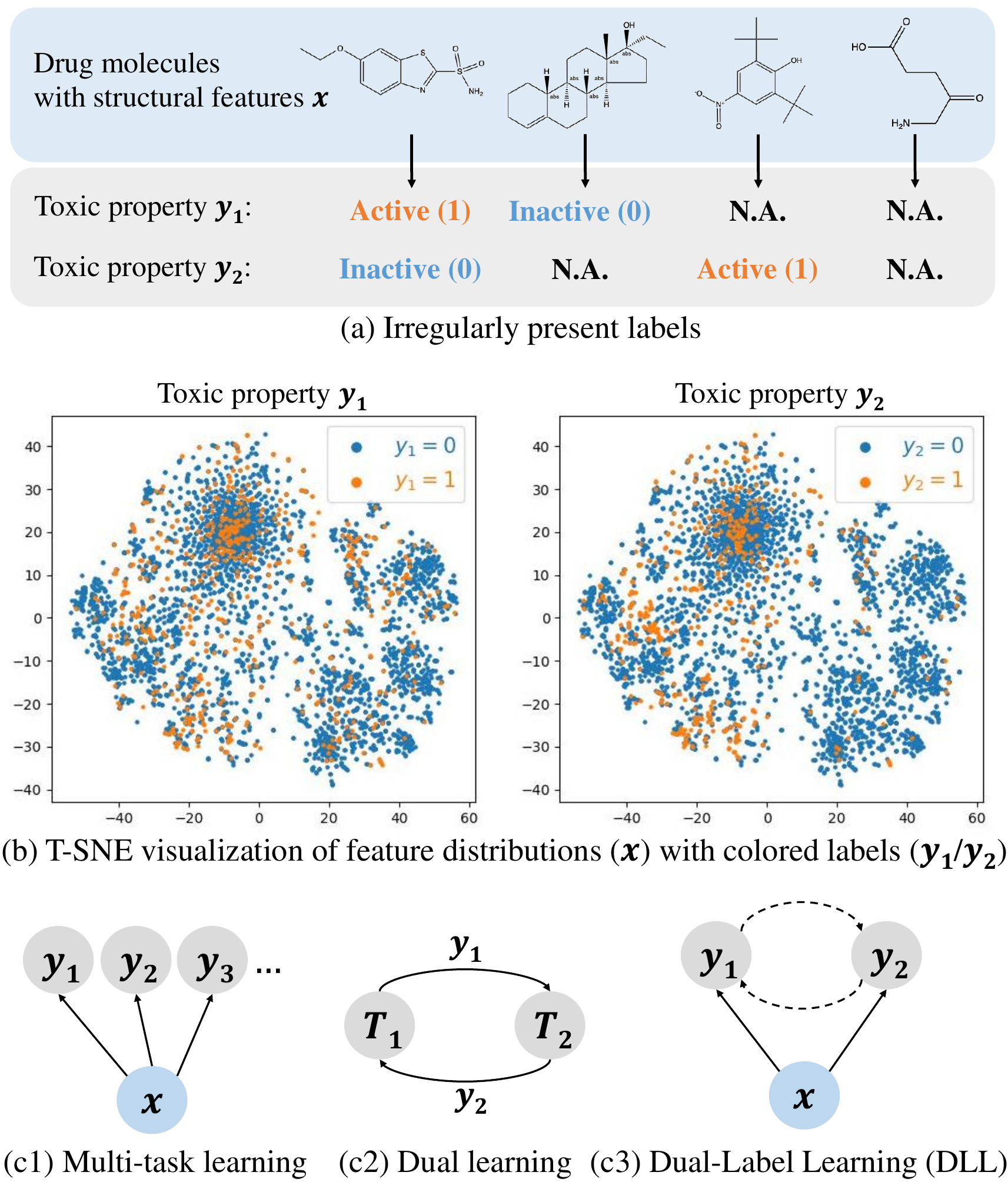}
\caption{Illustration of (a) irregularly present labels and (b) similar feature distributions of two labels with the Tox21 toxicity dataset. Labels $y_1$ and $y_2$ are two toxic properties of drug molecules in compound activity. `N.A.' means the label is missing. (c) Comparison of three learning mechanisms. $T_1$, $T_2$ denote two tasks.}
\label{fig:intro}
\end{figure}

In multi-task learning \cite{MTL}, labels across samples often exhibit irregular patterns. For instance, in drug analysis, multiple toxicity properties of a drug molecule might not be concurrently available due to experimental limitations. Figure \ref{fig:intro}(a) illustrates the irregular label presence in Tox21 \cite{tox21} toxicity dataset: some drug molecules are fully labeled, and others might have missing values for certain compound activities, leading to partially labeled (\textit{i.e. semi-labeled}) or totally \textit{unlabeled}  samples. Irregular label presence complicates the learning process and is common across various domains, including high-energy physics \cite{higgs}, material synthesis \cite{mof}, and even beyond scientific research, image annotation \cite{CV,cour2011learning} and music emotion classification \cite{music}.

Irregular label presence poses a significant challenge to machine learning tasks. Traditional multi-task learning methods \cite{MTL}, particularly multi-label learning \cite{MLL}, typically assume fully labeled samples during training, rendering them ineffective when handling missing labels. Although approaches have been proposed to leverage matrix completion or graph neural networks to address missing labels, they are largely confined to a specific type of multi-label classification tasks \cite{MLLsurvey}.
Even in recent advancements, neither large language models \cite{FunSearch,darwin,scbert,zhou2025bslora} nor foundation models \cite{GMAI,aurora} are applicable for tackling irregularly present labels in practice when large-scale data supports are unavailable. 
In practice, irregular label presence would deteriorate the model performance in making predictions.

The practical challenge necessitates a new learning mechanism capable of accommodating irregular labels and maximizing the use of any available label information from data. To meet this need, we envision a novel training and inference framework, which is tailored to handle any potential existence of missing labels and thoroughly learn from all available labels during the training process, and to fully utilize the known partial labels to infer the unknown labels during the inference process. The rationale behind this framework lies in the inherent correlations among labels. In Figure \ref{fig:intro}(b), the two toxicity properties in the Tox21 dataset exhibit similar feature distributions, indicating a strong correlation between the two. Therefore, inference can be achieved by explicitly modeling correlations among the labels, conditioned on the available features.

Conventional multi-task learning framework (Figure \ref{fig:intro}(c1)) models label correlations rather indirectly via shared feature inputs and share-bottom architecture among tasks, resulting in limited performance improvement \cite{MTL,MLL}.
To model label correlations more explicitly, we draw inspiration from dual learning to enhance multi-task learning performance. Originally designed for learning from unlabeled data, Dual Learning \cite{DL_book} leverages structural duality between tasks, where the output of one task serves as the input to another, creating a feedback loop (Figure \ref{fig:intro}(c2)). Similarly, we design a feedback loop for the two-label\footnote{Though designed for the two-label correlation, DLL applies to the general multi-label case via straightforward manipulation.} correlation
to maximize the utility of irregularly present labels and propose the novel Dual-Label Learning (DLL) framework (Figure \ref{fig:intro}(c3)). Unlike traditional dual learning that applies to two separate tasks with distinct models, DLL integrates the concept of a feedback loop into a single multi-task learning framework. Additionally, DLL conditions the joint label distribution on known features and specifically addresses samples with irregularly present labels, rather than unlabeled samples as in traditional dual learning.

The DLL framework formulates the problem as a dual-function system, where each function incorporates the output label of the other as part of its input, enabling direct information exchange between them. Therefore, the correlation between the two labels is explicitly modeled by the two functions in a \textit{bi-directional} manner. Mathematically, the feedback mechanism of the dual functions can be regarded as a \textit{bivariate} system of two equations, 
% (Section \ref{problem_definition})
solving which for the two unknowns would fulfill the inference. To estimate these functions from data, we propose a novel dual-tower model architecture designed to handle irregularly labeled training samples effectively.
% (Section \ref{dual_tower_model}).

\vspace{0.1 in}
Our main contributions are summarized as follows:

\begin{itemize}
\item We address the significant challenge of irregularly present labels in multi-task learning by introducing a novel Dual-Label Learning (DLL) framework. This framework uniquely formulates the problem into a dual-function system, wherein both functions are required to simultaneously satisfy standard supervision, structural duality and probabilistic duality (Section \ref{problem_definition}). Through rigorous theoretical analysis, we establish a generalization bound for DLL, grounded in Rademacher complexity, providing a robust theoretical foundation for its efficacy (Section \ref{theoretical_analysis}).
\item DLL introduces an innovative dual-tower model architecture, specifically designed to harness the interdependencies between labels. Such architecture allows for explicit information exchange between labels, effectively maximizing the utility of partially available labels and enhancing the understanding of label correlations (Section \ref{model_architecture}). Further, we propose a novel training scheme that incorporates label imputation directly into forward propagation (Section \ref{training}), and an inference scheme that jointly solves for missing labels as unknowns within a bivariate system of equations (Section \ref{inference}).
% \item We conduct extensive experiments to verify the effectiveness of explicitly modeling label correlation and maximizing the utility of available labels in DLL (Section \ref{experiment}). Results show that our method makes consistently better predictions than baseline approaches by up to a 10\% gain in F1-score or MAPE. Remarkably, our method provided with data at a label missing rate as high as 60\% can achieve similar or even better results than baseline approaches at a label missing rate of only 10\%. 
\item Extensive experiments demonstrate the superiority of DLL in various domains, showcasing its ability to outperform baselines  consistently. Our results indicate that DLL achieves up to a 10\% improvement in F1-score or MAPE, demonstrating its effectiveness in scenarios with high rates of missing labels. Notably, DLL maintains robust performance even when up to 60\% of the labels are missing, achieving results comparable to or better than baseline methods that operate under far less challenging conditions (Section \ref{experiment}). This highlights the framework's resilience and potential for broad application across different fields where data irregularities are prevalent.
\end{itemize}

\section{Dual-Label Learning (DLL) Framework} 
\subsection{Problem Formulation} \label{problem_definition}

\begin{figure}[t]
\centering
\includegraphics[width=0.98\columnwidth]{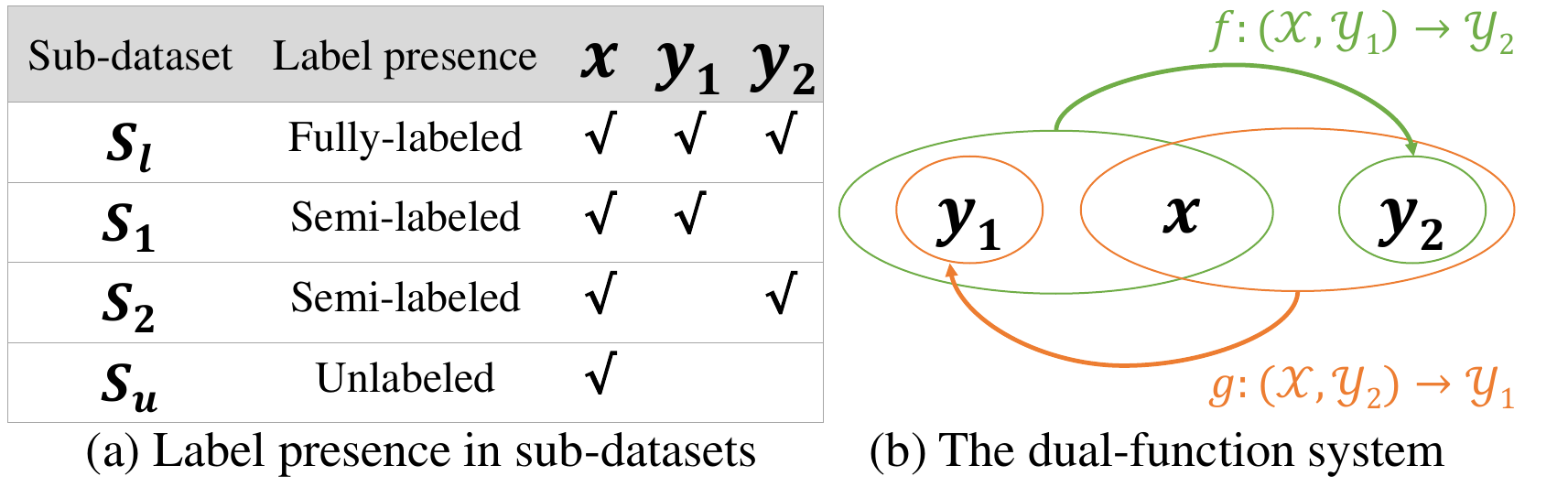}
\caption{Formulation of the Dual-Label Learning (DLL) framework. (a) The dataset split on different types of label presence and (b) the dual-function system in DLL.}
\label{fig:problem_definition}
\end{figure}

Consider a two-label learning problem where \textit{i)} labels are missing irregularly and independently and \textit{ii)} labels are correlated. 
More specifically, consider a dataset $\mathcal{S}={\{(x^{(i)}, y_1^{(i)}, y_2^{(i)})\}}_{i=1}^n$ drawn \textit{i.i.d.} from space $\mathcal{X} \times \mathcal{Y}_1 \times \mathcal{Y}_2$ according to an unknown underlying distribution $\mathcal{D}|_{\mathcal{X} \times \mathcal{Y}_1 \times \mathcal{Y}_2}$. $y_1^{(i)}$'s and/or $y_2^{(i)}$'s are irregularly present labels, missing completely at random. Therefore, samples in $\mathcal{S}$ can be divided into four sub-datasets: $\mathcal{S}_l$, $\mathcal{S}_1$, $\mathcal{S}_2$, and $\mathcal{S}_u$ as illustrated in Figure \ref{fig:problem_definition}(a). Let $I_l, I_1, I_2, I_u$ denote the index sets of the four sub-datasets respectively.

We propose a Dual-Label Learning (DLL) framework that formulates the two-label learning problem into a dual-function system, as depicted in Figure \ref{fig:problem_definition}(b). DLL aims to predict the two labels $y_1$ and $y_2$ simultaneously from features $x$ by finding the following two functions: (i) a \textit{primal} function $f: \mathcal{X} \times \mathcal{Y}_1 \rightarrow \mathcal{Y}_2$ to predict $y_2$ as close as possible to its ground truth; and (ii) a \textit{dual} function $g: \mathcal{X} \times \mathcal{Y}_2 \rightarrow \mathcal{Y}_1$ to predict $y_1$ as close as possible to its ground truth. The two models $f$ and $g$ are parameterized by $\theta_f$ and $\theta_g$, respectively:
\begin{equation} \label{objective12}
\begin{split}
f(x, y_1; \theta_f) := arg \mathop{max}\limits_{y_2' \in \mathcal{Y}_2} P(y_2'|x, y_1; \theta_f), \\
g(x, y_2; \theta_g) := arg \mathop{max}\limits_{y_1' \in \mathcal{Y}_1} P(y_1'|x, y_2; \theta_g),
\end{split}
\end{equation}
and are supervised by losses $l_1(g(x, y_2), y_1)$ and $l_2(f(x, y_1), y_2)$, respectively. The values of labels $y_1$ and $y_2$ in a particular sample can be inferred by solving a bivariate system of equations in a simple, symmetric format as follows:
\begin{equation} \label{equation_system}
\begin{cases}
y_2 = f(x, y_1; \theta_f), \\
y_1 = g(x, y_2; \theta_g).
\end{cases}
\end{equation}

To estimate the dual functions from the dataset $\mathcal{S}$, we propose a constrained multi-objective optimization problem for DLL to solve. The problem has four objectives and an equality constraint.

\begin{equation} \label{problem:DLL}
\begin{split}
\mathop{min}\limits_{\theta_f} \ & \frac{1}{|\mathcal{S}_l|} \sum_{i \in I_l} l_2 (f(x^{(i)}, y_1^{(i)}; \theta_f), y_2^{(i)}),\\
\mathop{min}\limits_{\theta_g} \ & \frac{1}{|\mathcal{S}_l|} \sum_{i \in I_l} l_1 (g(x^{(i)}, y_2^{(i)}; \theta_g), y_1^{(i)}),\\
\mathop{min}\limits_{\theta_f, \theta_g} \ & \frac{1}{|\mathcal{S}_l|+|\mathcal{S}_1|} \sum_{i \in I_l \cup I_1} l_1 (g(x^{(i)}, f(x^{(i)}, y_1^{(i)}; \theta_f); \theta_g), y_1^{(i)}), \\
\mathop{min}\limits_{\theta_f, \theta_g} \ & \frac{1}{|\mathcal{S}_l|+|\mathcal{S}_2|} \sum_{i \in I_l \cup I_2} l_2 (f(x^{(i)}, g(x^{(i)}, y_2^{(i)}; \theta_g); \theta_f), y_2^{(i)}),\\
s.t. \ & P(y_1|x)P(y_2|x,y_1;\theta_f) = P(y_2|x)P(y_1|x,y_2;\theta_g), \\
& \forall (x, y_1, y_2) \in \mathcal{X} \times \mathcal{Y}_1 \times \mathcal{Y}_2.
\end{split}
\end{equation}

The first two objectives are naturally derived from the two standard supervision loss functions $l_1$ and $l_2$ on $\mathcal{S}_l$.
% \begin{equation} \label{obj1}
% \mathop{min}\limits_{\theta_f} \frac{1}{|\mathcal{S}_l|} \sum_{i \in I_l} l_1 (f(x_i, y_{1,i}; \theta_f), y_{2,i}),
% \end{equation}
% \begin{equation} \label{obj2}
% \mathop{min}\limits_{\theta_g} \frac{1}{|\mathcal{S}_l|} \sum_{i \in I_l} l_2 (g(x_i, y_{2,i}; \theta_g), y_{1,i}).
% \end{equation}

The last two objectives take inspiration from the dual learning framework \cite{DL} and leverage the partially available labels in $\mathcal{S}_1$ and $\mathcal{S}_2$. They follow the structural duality of $f$ and $g$, aimed at minimizing the reconstruction losses in a two-agent communication game between $f$ and $g$.
% \begin{equation} \label{obj3}
% \mathop{min}\limits_{\theta_f, \theta_g} \frac{1}{|\mathcal{S}_l|+|\mathcal{S}_1|} \sum_{i \in I_l \cup I_1} l_2 (g(x_i, f(x_i, y_{1,i}; \theta_f); \theta_g), y_{1,i}), 
% \end{equation}
% \begin{equation} \label{obj4}
% \mathop{min}\limits_{\theta_f, \theta_g} \frac{1}{|\mathcal{S}_l|+|\mathcal{S}_2|} \sum_{i \in I_l \cup I_2} l_1 (f(x_i, g(x_i, y_{2,i}; \theta_g); \theta_f), y_{2,i}).
% \end{equation}
To illustrate the third objective, take a sample $(x^{(i)}, y_1^{(i)}) \in \mathcal{S}_1$. Then, a well-trained $f$ would map $(x^{(i)}, y_1^{(i)})$ to $\hat{y}_2^{(i)}$ close to the underlying ground truth $y_2^{(i)}$, and from $(x^{(i)}, \hat{y}_2^{(i)})$, a well-trained $g$ would reconstruct $\hat{y}_1^{(i)}$ close to the ground truth $y_1^{(i)}$. The same logic applies to the fourth objective.

The equality constraint reflects the probabilistic duality of $f$ and $g$, similar to \cite{DSL} and it holds according to the Bayes' Theorem.
% $f$ and $g$ should satisfy the following equality constraint,
% \begin{equation} \label{probabilistic_duality}
% \begin{split}
% P(y_1|x)P(y_2|x,y_1;\theta_f) = P(y_2|x)P(y_1|x,y_2;\theta_g), \\
% \forall x, y_1, y_2,
% \end{split}
% \end{equation}
The marginal probabilities $P(y_1|x)$ and $P(y_2|x)$ can be estimated via supervised learning on $\mathcal{S}_l \cup \mathcal{S}_1$ and $\mathcal{S}_l \cup \mathcal{S}_2$, respectively.

% To summarize, DLL aims to solve the constrained multi-objective optimization problem:
% \begin{equation} \label{DLL}
% \begin{split}
% \mathop{min} \limits_{\theta_f, \theta_g}  \ & \text{(\ref{obj1}), (\ref{obj2}), (\ref{obj3}), (\ref{obj4})} \\
% s.t. \ & \text{(\ref{probabilistic_duality})}.
% \end{split}
% \end{equation}

\subsection{Theoretical Analysis} \label{theoretical_analysis}

In this section, we prove a theoretical guarantee for DLL in terms of the generalization error bound based on Rademacher complexity.

For ease of notation, we first introduce an auxiliary vector $u \in \mathcal{U} = \mathbb{Z}_2^2$ to indicate which labels are known/missing in a sample $(x, y_1, y_2)$:
\begin{equation} \label{indicator}
u := (u_1, u_2) := (\mathbb{I}[y_1 \text{ is known}], \mathbb{I}[y_2 \text{ is known}]).
\end{equation}
For example, a sample $i \in I_l$ is associated with an indicator $u^{(i)} = (1, 1)$. By the randomness assumption, we have $u \perp (x, y_1, y_2)$.

Consider the space $\mathcal{X} \times \mathcal{Y}_1 \times \mathcal{Y}_2 \times \mathcal{U}$. Then the dataset $\mathcal{S}$ can been seen as drawn together with $\{u^{(i)}\}_{i=1}^n$ from a fixed distribution $\mathcal{D}|_{\mathcal{X} \times \mathcal{Y}_1 \times \mathcal{Y}_2 \times \mathcal{U}}$. Let $\mathcal{F} = \{ f(x, y_1; \theta_f); \theta_f \in \Theta_f \}$ be the primal function space, and $\mathcal{G} = \{ g(x, y_2; \theta_g); \theta_g \in \Theta_g \}$ be the dual function space, where $\Theta_f$ and $\Theta_g$ are parameter spaces.

Theoretically, DLL is to minimize the expected risk defined as
\begin{equation}
R(f, g) = \mathbb{E}_\mathcal{D}[l_{f, g}(x, y_1, y_2, u)], \forall (f, g) \in \mathcal{H},
\end{equation}
where the loss function $l_{f, g}: \mathcal{X} \times \mathcal{Y}_1 \times \mathcal{Y}_2 \times \mathcal{U} \rightarrow \mathbb{R}$ is
\begin{equation} \label{expected_risk}
\begin{split}
l_{f, g}(x, y_1, y_2, u) 
= & \ \alpha_1 \mathbb{I}[u=(1, 1)] \cdot l_2 (f(x, y_1), y_2) \\
+ & \ \alpha_2 \mathbb{I}[u=(1, 1)] \cdot l_1 (g(x, y_2), y_1) \\
+ & \ \alpha_3 \mathbb{I}[u_1=1] \cdot l_1 (g(x, f(x, y_1)), y_1) \\
+ & \ \alpha_4 \mathbb{I}[u_2=1] \cdot l_2 (f(x, g(x, y_2)), y_2),
\end{split}
\end{equation}
which is a linear combination of the objectives in (\ref{problem:DLL}) given an arbitrary set of coefficients $\alpha \in [0, 1]^4$ subject to $\sum_{j=1}^4 \alpha_j = 1$, and $\mathcal{H} \subseteq \mathcal{F} \times \mathcal{G}$ denotes the subspace of $(f, g)$ that satisfies the equality constraint in (\ref{problem:DLL}). For simplicity, we only consider loss functions $l_1: \mathcal{Y}_1 \times \mathcal{Y}_1 \rightarrow [0, 1]$ and $l_2: \mathcal{Y}_2 \times \mathcal{Y}_2 \rightarrow [0, 1]$, and it follows obviously at $l_{f, g}: \mathcal{X} \times \mathcal{Y}_1 \times \mathcal{Y}_2 \times \mathcal{U} \rightarrow [0, 1]$.

The corresponding empirical risk is defined on the $n$ samples as follows:
\begin{equation}
R_n(f, g) = \frac{1}{n} \sum_{i \in n} l_{f, g}(x^{(i)}, y_1^{(i)}, y_2^{(i)}, u^{(i)}), \forall (f, g) \in \mathcal{H}.
\end{equation}

Now we are ready to introduce Rademacher complexity \cite{rademacher}.

\begin{definition}[Rademacher complexity] \label{def:rademacher_complexity}
The Rademacher complexity of DLL is defined as
\begin{equation}
R_n^{DLL} = \mathbb{E}_{\mathcal{D},\sigma}[\mathop{sup}\limits_{(f,g) \in \mathcal{H}} \frac{1}{n} \sum_{i=1}^n \sigma^{(i)} \cdot l_{f, g}(x^{(i)}, y_1^{(i)}, y_2^{(i)}, u^{(i)})],
\end{equation}
where $\sigma^{(i)}$'s are i.i.d. variables uniformly drawn from $\{-1, 1\}$, known as Rademacher variables, and $\mathcal{D}$ is short for $\mathcal{D}|_{\mathcal{X} \times \mathcal{Y}_1 \times \mathcal{Y}_2 \times \mathcal{U}}$.
\end{definition}

Then according to \cite{rademacher}, we can derive a guarantee of the generalization error bound for DLL based on Rademacher complexity.

\begin{theorem} \label{theorem:DLL_bound}
For any $\delta > 0$, with probability at least $1 - \delta$, the following inequality holds for all $(f, g) \in \mathcal{H}$:
\begin{equation}
R(f, g) \leq R_n(f, g) + 2R_n^{DLL} + \sqrt{\frac{log \frac{1}{\delta}}{2n}}.
\end{equation}
\end{theorem}

\section{Learning with a Dual-Tower Model} \label{dual_tower_model}
\begin{figure*}
  \includegraphics[width=\textwidth]{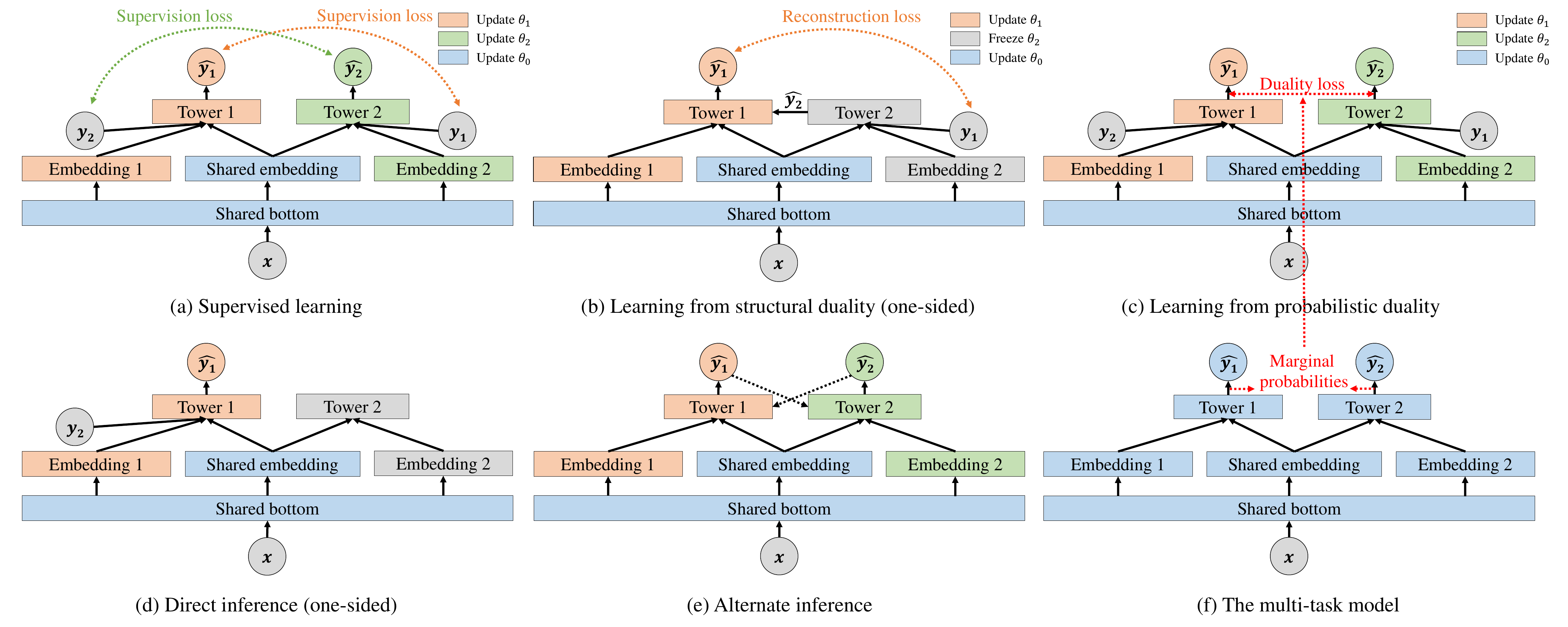}
  \caption{The dual-tower model (a-e) in DLL, its training (a-c) and inference (d-e); the multi-task model (f) to estimate the duality loss in (c).}
  \label{fig:dual_tower_model}
\end{figure*}

\subsection{Model architecture} \label{model_architecture}

We devise a novel dual-tower architecture to model the dual functions $f$ and $g$ in a single framework, as shown in Figure \ref{fig:dual_tower_model}(a-e). Here, $f$ is modeled by Tower 2 along with its parent nodes, while $g$ is modeled by Tower 1 along with its parent nodes. Note that $f$ and $g$ partially share parameters to model their common input $x$. To facilitate notation, for the rest of the paper, we denote the shared parameters separately by $\theta_0$ (in blue blocks), leaving the rest of the parameters in $f$ as $\theta_2$ (in green blocks) and the rest of the parameters in $g$ as $\theta_1$ (in orange blocks). Duality manifests itself on the top of the architecture, where the output $y_2$ of $f$ serves as an input in $g$, while the output $y_1$ of $g$ serves as an input in $f$. It ties with the symmetric equation system in (\ref{equation_system}).

Within the framework, configurations of the two towers and their parent nodes are flexible, subject to the task (\textit{e.g.} classification or regression), the input data modality (\textit{e.g.} images, text, or graphs), and the choice of building blocks (\textit{e.g.} attention, fully connected layer, or mixture of experts).

% \begin{figure}[t]
% \centering
% \includegraphics[width=0.98\columnwidth]{multitask_model.pdf}
% \caption{The multi-task model to estimate the marginal probability.}
% \label{fig:multitask_model}
% \end{figure}

\subsection{Training} \label{training}

We propose a novel training algorithm of the dual-tower model to solve the multi-objective constrained optimization problem in DLL for the parameters $\theta_{0}, \theta_{1}, \theta_{2}$. The training algorithm is illustrated in Figure \ref{fig:dual_tower_model}(a-c) and detailed in Algorithm \ref{algo:DLL_train}. The core idea is to integrate training modes (a-c) to flexibly accommodate irregularly present labels.

Figure \ref{fig:dual_tower_model}(a) illustrates supervised learning to minimize the supervision losses for fully labeled samples in $\mathcal{S}_l$.

Figure \ref{fig:dual_tower_model}(b) illustrates learning from structural duality by minimizing the reconstruction loss for samples in $\mathcal{S}_1$ where only label $y_1$ is available. In practice, we freeze Tower 2 (and hence Embedding 2) as there is no supervision on its direct output $\hat{y_2}$. This avoids the two towers from cheating on a simple 'copy-paste' task. The training for the other tower (with samples in $\mathcal{S}_2$) only differs by symmetry.

Figure \ref{fig:dual_tower_model}(c) illustrates learning from probabilistic duality to fulfill the equality constraint in (\ref{problem:DLL}). To elaborate, the equality constraint is first relaxed and converted into a regularization term as part of the objective function, named as the `duality loss': $l_{d} (x, y_1, y_2; \theta_{0}, \theta_{1}, \theta_{2})   
= (\log P_M(y_1|x) + \log P_f(y_2|x, y_1; \theta_2, \theta_{0})
- \log P_M(y_2|x) - \log P_g(y_1|x, y_2; \theta_1, \theta_{0})) ^2.$
% \begin{equation}
% \begin{split}
% & l_{d} (x, y_1, y_2; \theta_{0}, \theta_{1}, \theta_{2}) \\   
% = & (\log P_M(y_1|x) + \log P_f(y_2|x, y_1; \theta_2, \theta_{0}) \\
% & - \log P_M(y_2|x) - \log P_g(y_1|x, y_2; \theta_1, \theta_{0})) ^2.
% \end{split}
% \end{equation}
Then, as shown in Figure \ref{fig:dual_tower_model}(c), the duality loss is exerted on the softmax layers of the two towers prior to the outputs, in which the marginal probabilities $P_M(y_1|x)$ and $P_M(y_2|x)$ are estimated with a pre-trained multi-task model $M$ as designed in Figure \ref{fig:dual_tower_model}(f). Finally, the dual-tower model, though trainable on the entire sample space, is trained with fully and partially labeled samples to minimize the duality loss.

Algorithm \ref{algo:DLL_train} organically integrates the three training modes (a-c) based on sample-wise label presence. Lines \ref{algo:impute_I1}-\ref{algo:impute_I2} impute missing labels for partially-labeled samples. Lines \ref{algo:loss_s1}-\ref{algo:loss_d} calculate the three types of losses. Lines \ref{algo:grad_G0}-\ref{algo:grad_G2} calculate the gradients. Lines \ref{algo:update_theta0}-\ref{algo:update_theta2} update the parameters. The computational complexity of Algorithm \ref{algo:DLL_train} is $O(knN)$ where $k$ refers to the complexity of a single feed-forward pass through the dual-tower model. A separate algorithm of each training mode (a-c, f) is available in Appendix \ref{appendix:DLL_train_modes}.

\begin{algorithm}[t]
\caption{The DLL Training Algorithm} \label{algo:DLL_train}

\begin{flushleft}
\textbf{Input:} 
Training datasets $\mathcal{S}_l, \mathcal{S}_1, \mathcal{S}_2$ with index sets $I_l, I_1, I_2$; supervision loss coefficients $\lambda_{12}, \lambda_{21}$; reconstruction loss coefficients $\lambda_{11}, \lambda_{22}$; duality loss coefficients $\lambda_{d}$; optimizer $Opt$; number of epochs $N$; batch size $B$

\textbf{Output:} 
the dual-tower model parameters $\hat{\theta}_{0}, \hat{\theta}_{1}, \hat{\theta}_{2}$
\end{flushleft}

\begin{algorithmic}[1]

\STATE Initialize $\hat{\theta}_{0}, \hat{\theta}_{1}, \hat{\theta}_{2}$;
\STATE \textbf{for} $epoch = 1, 2, ..., N$ \textbf{do} 
\STATE \hspace{0.5cm} \textbf{for} $batch$ with $B$ indices $\{i\} \subset I_l \cup I_1 \cup I_2$ \textbf{do} 
% \STATE \hspace{0.5cm}$\{(x^{(i)}, y_1^{(i)}, y_2^{(i)})\}_{i=1}^B$ from $\mathcal{S}_l \cup \mathcal{S}_1 \cup \mathcal{S}_2$;
\STATE \hspace{1cm} \textbf{for} $i \in I_1$  \textbf{do} $\hat{y}_2^{(i)} = f(x^{(i)}, y_1^{(i)}; \hat{\theta}_{2}, \hat{\theta}_{0})$ \textbf{end for}
\label{algo:impute_I1}
\STATE \hspace{1cm} \textbf{for} $i \in I_2$ \textbf{do}  $\hat{y}_1^{(i)} = g(x^{(i)}, y_2^{(i)}; \hat{\theta}_{1}, \hat{\theta}_{0})$ \textbf{end for} \label{algo:impute_I2}

% \STATE calculate loss
\STATE \hspace{1cm} $s_1 = \lambda_{21} \sum_{i \in I_l} l_1 (g(x^{(i)}, y_2^{(i)}; \hat{\theta}_{1}, \hat{\theta}_{0}), y_1^{(i)})$; \label{algo:loss_s1}
\STATE \hspace{1cm} $s_2 = \lambda_{12} \sum_{i \in I_l} l_2 (f(x^{(i)}, y_1^{(i)}; \hat{\theta}_{2}, \hat{\theta}_{0}), y_2^{(i)})$; \label{algo:loss_s2}
\STATE \hspace{1cm} $r_1 = \lambda_{11} \sum_{i \in I_1}
 l_1 (g(x^{(i)}, \hat{y}_2^{(i)}; \hat{\theta}_{1}, \hat{\theta}_{0}), y_1^{(i)})$; \label{algo:loss_r1}
 \STATE \hspace{1cm} $r_2 = \lambda_{22} \sum_{i \in I_2}
 l_2 (f(x^{(i)}, \hat{y}_1^{(i)}; \hat{\theta}_{2}, \hat{\theta}_{0}), y_2^{(i)})$; \label{algo:loss_r2}
 
 \STATE \hspace{1cm} $d = \lambda_{d} [ \sum_{i \in I_l} l_d (x^{(i)}, y_1^{(i)}, y_2^{(i)}; \hat{\theta}_0, \hat{\theta}_1, \hat{\theta}_2)$
 \STATE \hspace{2cm} $+ \sum_{i \in I_1} l_d (x^{(i)}, y_1^{(i)}, \hat{y}_2^{(i)}; \hat{\theta}_0, \hat{\theta}_1, \hat{\theta}_2)$
 \STATE \hspace{2cm} $+  \sum_{i \in I_2} l_d (x^{(i)}, \hat{y}_1^{(i)}, y_2^{(i)}; \hat{\theta}_0, \hat{\theta}_1, \hat{\theta}_2) ]$; \label{algo:loss_d}
 % \STATE calculate gradients
 \STATE \hspace{1cm} $G_0 = \nabla_{\theta_{0}} \frac{1}{B}(s_1+s_2+r_1+r_2+d)$; \label{algo:grad_G0}
 \STATE \hspace{1cm} $G_1 = \nabla_{\theta_{1}} \frac{1}{B}(s_1+r_1+d)$; \label{algo:grad_G1}
 \STATE \hspace{1cm} $G_2 = \nabla_{\theta_{2}} \frac{1}{B}(s_2+r_2+d)$; \label{algo:grad_G2}
 % \STATE update parameters:
 \STATE \hspace{1cm} $\hat{\theta}_{0} \leftarrow Opt(\hat{\theta}_{0}, G_0)$; \label{algo:update_theta0}
 \STATE \hspace{1cm} $\hat{\theta}_{1} \leftarrow Opt(\hat{\theta}_{1}, G_1)$; \label{algo:update_theta1}
 \STATE \hspace{1cm} $\hat{\theta}_{2} \leftarrow Opt(\hat{\theta}_{2}, G_2)$; \label{algo:update_theta2}
 \STATE \hspace{0.5cm} \textbf{end for}
 \STATE \textbf{end for}
 \STATE \textbf{return} $\hat{\theta}_{0}, \hat{\theta}_{1}, \hat{\theta}_{2}$

\end{algorithmic}
\end{algorithm}

\subsection{Inference} \label{inference}
We propose a novel inference algorithm for the learnt dual-tower model to predict missing labels in the dataset. The inference algorithm is illustrated in Figure \ref{fig:dual_tower_model}(d-e) and detailed in Algorithm \ref{algo:DLL_infer}.

Figure \ref{fig:dual_tower_model}(d) illustrates direct inference on a semi-labeled sample. The single unknown label $y_1$ (or $y_2$) is predicted from the other label with $g$ (or $f$) (Algorithm \ref{algo:DLL_infer} Lines \ref{algo:pred_I1}-\ref{algo:pred_I2}). 

Figure \ref{fig:dual_tower_model}(e) illustrates alternate inference on an unlabeled sample. Labels $y_1$ and $y_2$ become unknowns of a bivariate equation system (\ref{equation_system}). To solve for $y_1$ and $y_2$, Algorithm \ref{algo:DLL_infer} conducts alternate inference between $f$ and $g$ which, upon a proper label initialization (Line \ref{algo:alter_init}), repeatedly reuses the output labels from the other towers as inputs (Lines \ref{algo:alter_f}-\ref{algo:alter_g}), until they reach an equilibrium where the input labels are close to the output labels from the other towers (Line \ref{algo:alter_converge}).

Algorithm \ref{algo:DLL_infer} assumes that equilibrium is attainable and operates under carefully selected conditions to facilitate convergence. For example, in classification tasks, we use probabilistic outputs instead of binary labels within the for loop to enhance the likelihood of convergence. The computational complexity of Algorithm \ref{algo:DLL_infer} is $O(kL)$.

\begin{algorithm}[t]
\caption{The DLL Inference Algorithm} \label{algo:DLL_infer}

\begin{flushleft}
\noindent \textbf{Input:} 
A sample with index $i \in I_1 \cup I_2 \cup I_u$; estimated parameters $\hat{\theta}_{0}, \hat{\theta}_{1}, \hat{\theta}_{2}$; label initialization $y_0$; iterations $L$ 

\textbf{Output:} 
Predicted labels $\hat{y}_1^{(i)} \in \mathcal{Y}_1$ and/or $\hat{y}_2^{(i)} \in \mathcal{Y}_2$
\end{flushleft}

\begin{algorithmic}[1]

% \STATE (d) Direct inference: 
\STATE \textbf{if} $i \in I_1$ \textbf{then} $\hat{y}_2^{(i)} = f(x^{(i)}, y_1^{(i)}; \hat{\theta}_{2}, \hat{\theta}_{0})$ \textbf{return} $\hat{y}_2^{(i)}$ \label{algo:pred_I1}
\STATE \textbf{if} $i \in I_2$ \textbf{then} $\hat{y}_1^{(i)} = g(x^{(i)}, y_2^{(i)} \hat{\theta}_{1}, \hat{\theta}_{0})$ \textbf{return} $\hat{y}_1^{(i)}$ \label{algo:pred_I2}

% \STATE (e) Alternate inference: 
\STATE \textbf{if} $i \in I_u$ \textbf{then}
\STATE \hspace{0.5cm} Initialize $\hat{y}_{1}^{(i)} = \hat{y}_{2}^{(i)} = y_0$; \label{algo:alter_init}
\STATE \hspace{0.5cm} \textbf{for} $iteration = 1, 2, ..., L$ \textbf{do} 
\STATE \hspace{1cm} $\hat{y}_{2}^{(i)} = f(x^{(i)}, \hat{y}_{1}^{(i)}; \hat{\theta}_{2}, \hat{\theta}_{0})$; \label{algo:alter_f}
\STATE \hspace{1cm} $\hat{y}_{1}^{(i)} = g(x^{(i)}, \hat{y}_{2}^{(i)}; \hat{\theta}_{1}, \hat{\theta}_{0})$; \label{algo:alter_g}
\STATE \hspace{1cm} \textbf{if} $(\hat{y}_{1}, \hat{y}_{2})$ converge \textbf{then break} \label{algo:alter_converge}
\STATE \hspace{0.5cm} \textbf{end for} 
\STATE \hspace{0.5cm} \textbf{return} $\hat{y}_1^{(i)}, \hat{y}_2^{(i)}$

\end{algorithmic}
\end{algorithm}

\section{Experiments} \label{experiment}
We conduct extensive experiments on both real-world and synthetic datasets from various domains to evaluate the effectiveness of the DLL framework, entailing the dual-tower model and its training and inference algorithms.

\subsection{Datasets} \label{dataset}

Our experiments are conducted on three dataset: Tox21, HIGGS and MOF. 

\subsubsection{Tox21}
The Tox21 dataset is a real-world toxicity dataset from the 2014 Tox21 data challenge \cite{deeptox,tox21} to predict chemical compounds' interference in biochemical pathways using only chemical structure data. The raw data come from the the Toxicology in the 21st Century (Tox21) program, and can be downloaded from
\footnote{\url{http://bioinf.jku.at/research/DeepTox/tox21.html}}. 
For experiment, we use 7,831 chemical compounds from the dataset along with their chemical structures (represented in ``SMILES'') as features and two binary labels (``SR-ARE'' and ``SR-MMP'') related to toxicity. Both labels are partially missing at a rate of 26\%, and missing samples partially overlap. The two labels are correlated as both are related to the chemical compounds' toxicity, but the exact relationship is unclear. This suffices the setting for the DLL problem with a binary classification task.

\subsubsection{HIGGS}
The HIGGS dataset is a synthetic dataset to facilitate the study of searching for exotic particles in high-energy physics \cite{higgs}. The dataset contains 11,000,000 instances, each associated with a binary variable indicating Higgs bosons, 21 real variables related to kinetic properties, and 7 continuous high-level variables generated as functions of the 21 features. The dataset can be downloaded from \footnote{https://archive.ics.uci.edu/dataset/280/higgs}.
For experiment, we randomly select 10,000 instances, including the first 22 real variables as features and the first 2 out of 7 synthetic variables as dual labels. Therefore, the DLL problem is associated with a regression task. The dataset can be randomly split into the four label missing types according to pre-set missing rates.

\subsubsection{MOF}
The MOF dataset is a real-world dataset from Tianchi Competition for prediction of the synthesis conditions of metal-organic framework (MOF) materials \cite{mof}. The raw data can be downloaded from \footnote{https://tianchi.aliyun.com/competition/entrance/532116}. For experiment, we select the structure information of 729 MOF materials as well as their synthesis conditions. The temperature and time required to synthesize the material are two correlated condition variables and are selected as dual labels. Features are extracted from the structure information of MOF. Hence, DLL is associated with regression tasks.

% Experiments are conducted on three dataset: Tox21, HIGGS and MOF. The Tox21 dataset \cite{tox21} is a real-world dataset that entails the toxicity properties of 7,831 chemical compounds for drug analysis. The HIGGS dataset \cite{higgs} is a synthetic dataset that entails the kinetic properties of 10,000 particles in high-energy physic. The MOF dataset \cite{mof} is a real-world dataset that entails the structure information and synthesis conditions of 729 metal-organic framework materials. Table \ref{tab:meta_data} summarizes the meta data of the three datasets, including the split by label presence type, degree of authenticity, the corresponding tasks and evaluation metrics. Full details of the datasets are available in the Supplementary Materials.

Table \ref{tab:meta_data} summarizes the meta data of the three datasets. 
\begin{table}[htb] % [ht]
    \centering
      \begin{tabular}{l|r|r|r}
    \midrule
    \textbf{Dataset}&\textbf{Tox21}&\textbf{HIGGS}&\textbf{MOF}\\ 
    \midrule
    Total size	&7,831	&10,000	&729	\\
    $\mathcal{S}_l$ size	&4,460	&3,600	&262	\\
    $\mathcal{S}_1$ size	&1,372	&2,400	&175	\\
    $\mathcal{S}_2$ size	&1,350	&2,400	&175	\\
    $\mathcal{S}_u$ size	&649	&1,600	&117	\\
    \midrule
    Features        &128, Real   &7, Real       &166, Real\\
    Labels          &2, Real   &2, Synthetic  &2, Real\\
    Label $y_1$ &SR-ARE &- &Temperature \\
    Label $y_2$ &SR-MMP &- &Time \\
    Missing rates    &Real   &Synthetic*  &Synthetic*\\
    \midrule
    Task            &Binary classification &Regression & Regression \\
    Metrics & F1-score &MAPE & MAPE \\
    \midrule
    \multicolumn{4}{l}{\footnotesize{*By default, the missing rate is set as 40\% for each label.}}
  \end{tabular}
  \caption{Meta data of the three datasets.}
  \label{tab:meta_data}
\end{table}

\subsection{Baseline approaches}
Following \cite{DSML}, DLL is compared against the most comparable learning schemes, including:
% Independent Decomposition (ID) that ignores label correlation, Co-Occurrence Learning (COL) and Semi-Supervised Learning (SSL) that indirectly model label correlation, and Label Stacking (LS) and Dual Set Multi-Label Learning (DSML) that directly model label correlation, but in limited label communication stages. Details of baseline approaches are available in Appendix \ref{appendix:experiment}.
\begin{enumerate}
    \item \textbf{Independent Decomposition (ID)} \cite{DSML}. The DLL problem is decomposed into two sub-problems. Two separate models are learnt to estimate labels $y_1$ and $y_2$ independently: $M_1 \sim P(y_1|x)$, $M_2 \sim P(y_2|x)$. This approach ignores label correlation between $y_1$ and $y_2$.
    \item \textbf{Co-Occurrence Learning (COL)} \cite{DSML}. The DLL problem is regarded as a multi-output prediction problem. The model learns to estimate labels $y_1$ and $y_2$ jointly: $M \sim P((y_1, y_2) | x)$. Label correlation is learnt rather indirectly from fully labeled subset $\mathcal{S}_l$ only.
    \item \textbf{Semi-Supervised Learning (SSL)} \cite{DSML}. The DLL problem is decomposed into a two-stage multi-output prediction problem. At the first stage, a multi-output model is learnt: $M_1 \sim P((y_1, y_2)| x)$ from $\mathcal{S}_l$, and used to predict and impute the missing labels in $\mathcal{S}_1$ and $\mathcal{S}_2$; at the second stage, a new multi-output model is learnt $M_2 \sim P((y_1, y_2)| x)$ from the imputed $\mathcal{S}_l \cup \mathcal{S}_1 \cup\mathcal{S}_2$, and used to predict the labels for $\mathcal{S}_u$. This approach goes one step beyond Semi-Supervised Learning to leverage the partially labeled data, but still learns the label correlation indirectly.
    \item \textbf{Label Stacking (LS)} \cite{DSML}. The DLL problem is decomposed into two single-label learning problems, one after another. A first model is learnt to predict one label: $M_1 \sim P(y_2 | x)$ from $\mathcal{S}_l \cup \mathcal{S}_2$, and used to predict and impute the missing labels in $\mathcal{S}_1$; and a second model is learnt to predict the other label based on the output of the first model: $M_2 \sim P(y_1 | x, \hat{y}_2)$ from $\mathcal{S}_l \cup \mathcal{S}_1$. In this approach, label correlation is learnt in a single direction from $y_2$ to $y_1$. 
    \item \textbf{Dual Set Multi-Label Learning (DSML)} \cite{DSML}. We choose the model-reuse mechanism in DSML\footnote{For a fair comparison, we replace the boosting process with supervised learning on all available data.}, which goes one step beyond Label Stacking to learn the bi-directional label correlation. Step by step, a first model $M_1 \sim P(y_2|x)$ is learnt from $\mathcal{S}_l \cup \mathcal{S}_2$ to predict $\hat{y}_2$ in $\mathcal{S}_1$; then a second model $M_2 \sim P(y_1|x, \hat{y}_2)$ is learnt from $\mathcal{S}_l \cup \mathcal{S}_1$ to predict $\hat{y}_1$ in $\mathcal{S}_2$; finally, a third model $M_3 \sim P(y_2|x, \hat{y}_1)$ is learnt from $\mathcal{S}_l \cup \mathcal{S}_1 \cup\mathcal{S}_2$ to predict $y_2$ label. In this approach, label correlation is learnt bi-directionally, but in limited label communication stages.
    % \item \textbf{Multi-Task Learning}. The multi-task model in Figure \ref{fig:dual_tower_model}(f) naturally serves as a baseline for DLL, without the duality constraints: $M \sim P(y|x), P(z|x)$. The correlation between the two labels are learnt indirectly via training on the shared bottom in the model architecture.
    % \item first step 2 iter, then step 3 iter
    % \item iter: step 2 (iter) <-> step 3 (iter)
    % \item freeze vs not freeze
\end{enumerate}

\subsection{Experimental settings} \label{settings}

Prediction tasks fall into two types subject to label presence: single-label prediction for semi-labeled samples, of which only one label is to be predicted based on the other label available; and double-label prediction for unlabeled samples, of which both labels are unknown and to be predicted. By notation, for example, ``Single-$y1$'' refers to prediction of label $y1$ in single-label prediction, where label $y2$ is known and label $y1$ can be obtained via direct inference. From a different perspective, tasks fall into two types subject to label variable type, including classification and regression.

Evaluation metrics primarily include the F1-score for binary classification tasks on the Tox21 dataset, where a higher F1-score indicates better performance; and the Mean Absolute Percentage Error (MAPE) for regression tasks on HIGGS and MOF, where a lower MAPE indicates better performance.

Detailed settings of the hardware, data-preprocessing, the model architecture, training and inference are available in the Supplementary Materials. 

\subsection{Results and analysis}

\subsubsection{Comparison to baselines}
\label{exp:baseline}

\begin{figure*}
  \includegraphics[width=\textwidth]{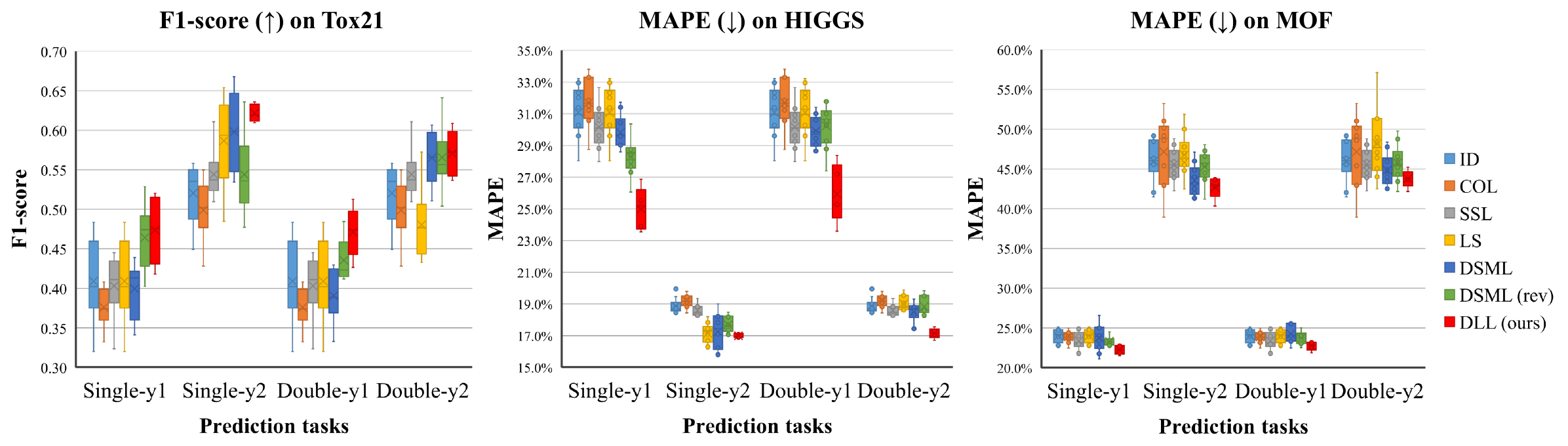}
  \caption{Box plots of baseline comparison on three datasets. Each dataset is associated with four prediction tasks. DLL applies direct inference to single-label prediction (`Single') and alternate inference to double-label prediction (`Double').  Each box summarizes the results of a single method executing 10 times. Exact numbers are available in Appendix \ref{appendix:exp_result}. Baseline approaches are detailed in Appendix \ref{appendix:baseline}.}
  \label{fig:baseline}
\end{figure*}

DLL is compared with various baseline approaches on both synthetic and real-world datasets for classification and regression tasks. Results are summarized in Figure \ref{fig:baseline}.
Overall, our approach DLL (in red boxes) consistently outperforms baselines across the three datasets in terms of average F1-score or MAPE. It can achieve up to 13\% performance improvement over the best-performing baseline in certain tasks (e.g. Double-$y1$ on HIGGS). This validates the performance gain of DLL's explicit modeling of bi-directional label correlation over single-round bi-directional label correlation modeling (DSML), uni-directional label correlation modeling (LS), indirect label correlation modeling (SSL, COL) and label-correlation-absent modeling (ID).

Task-wise, all approaches consistently produce better or equal results in single-label prediction, when the other label is known, than in double-label prediction. It aligns with our assumption that partially available labels carry non-negligible information, which should be exploited to the most extent during inference.

\subsubsection{Sensitivity to label missing rate}

\begin{figure*}[t]
\centering
\includegraphics[width=2.1\columnwidth]{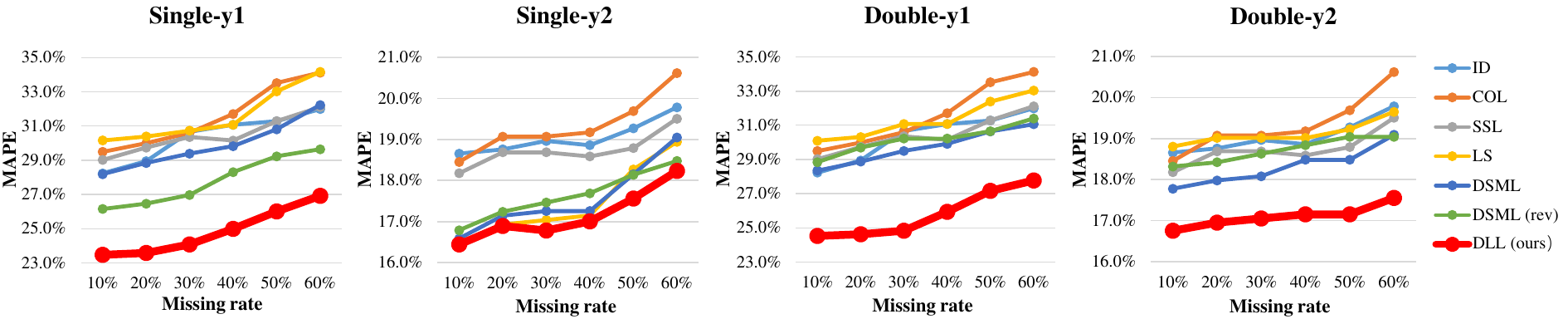}
\caption{MAPE trend over varied missing rates from 10\% to 60\% on HIGGS dataset. Our method is bolded in red.}
\label{fig:sensitivity}
\end{figure*}

We analyze the robustness of DLL as the label missing rate increases on the synthetic HIGGS dataset. The label missing rate is varied from 10\% to 60\% at a 10\% step size by randomly and independently masking label $y_1$ and label $y_2$ among samples at respective ratios. Figure \ref{fig:sensitivity} visualises the performance trends of DLL and baseline approaches over the increasing label missing rate. 

A few observations are drawn from Figure \ref{fig:sensitivity}. 
First, DLL maintains superior to baselines at different label missing rates, across datasets and tasks. MAPE results roughly divide the approaches into three tiers: tier 1 being DLL which explicitly models label correlation via the feedback loop, tier 2 being DSML and LS with limited label communication rounds, and tier 3 being SSL, COL and ID without direct label correlation modeling. The tiering is in line with our observation in Figure \ref{fig:baseline}. 
Second, increasing label missing rate causes MAPE to increase, worsening the performance of all approaches in general, which meets our expectation. Finally and remarkably, the superiority of DLL is rather significant in double-label prediction, where DLL operating at a missing rate of as high as 60\% can achieve even better results than baseline approaches operating at a far lower missing rate of only 10\%. This observation reaffirms the effectiveness of DLL in maximizing the utility of available labels by directly modeling the label correlation, training with label imputation, and solving the bi-variate equation system through alternate inference.

\subsubsection{Ablation study}

\begin{figure}[t]
  \includegraphics[width=0.98\columnwidth]{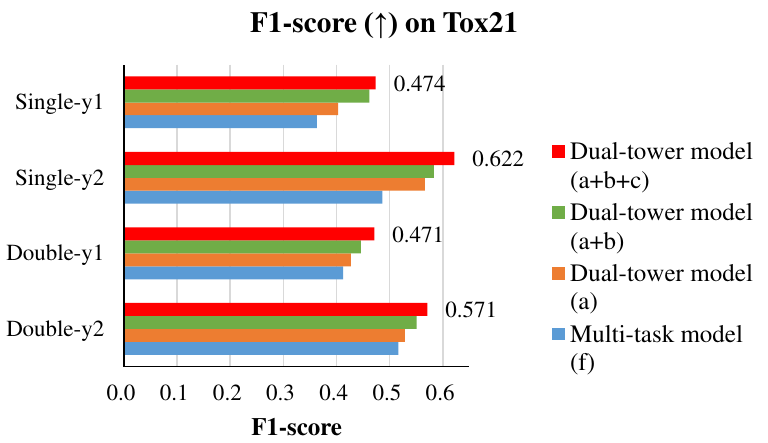}
  \caption{Ablation study of the dual-tower model and its training mechanism on Tox21 dataset. Our original method is colored in red.}
  \label{fig:ablation}
\end{figure}

We examine the necessity of each component in DLL, including the model architecture and training algorithms. The dual-tower model architecture is compared against (f) multi-task model, and the three training modes, including (a) supervised learning, (b) learning from structural duality and (c) learning from probabilistic duality, are added and tested sequentially. The experiments are conducted on the Tox21 dataset, and results are summarized in Figure \ref{fig:ablation}. Results show that the outstanding performance of DLL benefits from the design of each of its components.

% \begin{table}[t]

% \setlength{\tabcolsep}{1.2mm}
%   \begin{tabular}{l|cc|cc}
%     \midrule
%     { } & \multicolumn{2}{c|}{\textbf{Single-label}} & \multicolumn{2}{c}{\textbf{Double-label}}\\ 
%     \textbf{Method} & \multicolumn{2}{c|}{\textbf{prediction}} & \multicolumn{2}{c}{\textbf{prediction}}\\ 
%     \cmidrule{2-5} &$y$&$z$&$y$&$z$\\ 
%     \midrule
%     Multi-task model (f)	&0.395	&0.491	&0.395	&0.491\\
% Dual-tower model (a)	&0.438	&0.572	&0.409	&0.503\\
% Dual-tower model (a+b)	&0.502	&0.589	&0.427	&0.524\\
%     \midrule
%     \textbf{Dual-tower model (a+b+c)}	& \textbf{0.515}	&\textbf{0.628} &\textbf{0.451}	&\textbf{0.543} \\
%     \midrule
%   \end{tabular}
%   \caption{Ablation study of the dual-tower model and its training mechanism on the Tox21 dataset.}
%   \label{tab:ablation}
% \end{table}

\subsubsection{Convergence of alternate inference}

\begin{figure}[t]
  \includegraphics[width=0.98\columnwidth]{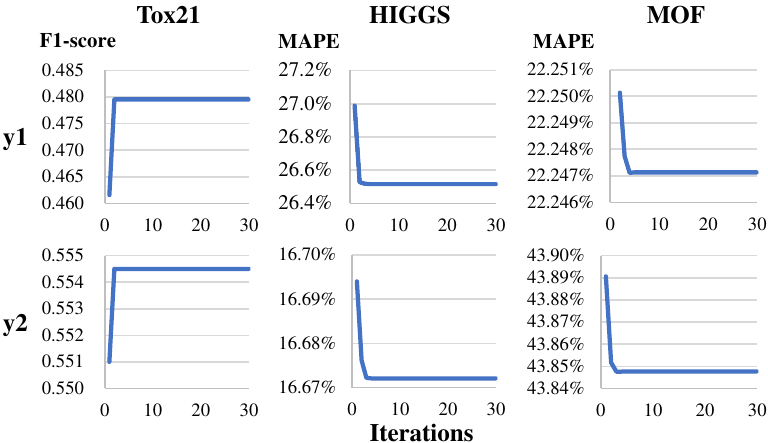}
  \caption{Performance trend against the number of iterations in alternate inference for double-label prediction.}
  \label{fig:convergence}
\end{figure}

In our experiments, alternative inference converges in the double-label prediction tasks on all the three datasets. We further analyze the convergence rate of alternative inference by plotting the performance trend against the number of iterations in Figure \ref{fig:convergence}. As the figure shows, alternate inferences usually converges fast, not beyond 10 iterations, though our iteration hyperparameter is set as $L=1,000$ in experiments. Despite of the iterations, the time cost of alternative inference is low: less than 10 seconds for executing 1,000 iterations to estimate labels of 2,000 samples in a dataset.

\section{Related Work}
\subsection{Multi-task Learning and Multi-label Learning}
Multi-task learning \cite{MTL, li2025unimatch} is a learning scheme that leverages useful information from multiple related tasks. DLL is a novel framework proposed for multi-task learning particularly with irregularly present labels. In DLL, the dual-tower model can be seen as an adapted version of Multi-gate Mixture-of-Experts (MMoE) \cite{MMoE}, a popular multi-task learning approach. MMoE features multiple expert networks which are selectively activated during training and inference by a gating mechanism to serve different tasks, while the dual-tower model replaces the gating mechanism with output labels from the other tower\footnote{This is equivalent to an implicitly hard-coded 0-1 gate.} to facilitate the special treatment for missing labels.

Multi-label learning \cite{MLL,MLLsurvey,class_imbalance_MLL,MLL2} is a special case of multi-task learning. The term is typically reserved for the context of classification in computer vision where multiple binary labels are assigned to the same image\cite{dual_relation_learning}, and tasks often share the same input space and feature representation. In contrast, DLL caters to both classification and regression tasks, and it features a feedback mechanism among tasks.
Among multi-label learning approaches, Dual Set Multi-Label Learning (DSML) \cite{DSML} is the closest to DLL in terms of problem definition, but DSML is limited to binary classification tasks and fails to accommodate missing labels.

\subsection{Dual Learning}
The dual learning paradigm is originally proposed in machine translation \cite{DL}, which introduces the structural duality between two tasks to leverage unlabeled data. Additionally, Dual Supervised Learning \cite{DSL} introduces the probabilistic duality as a regularization term, and Dual Inference \cite{DI} leverages structural duality during the inference stage, combining the loss functions of dual tasks to improve inference. 
Dual learning has been connected to various domains including transfer learning \cite{DTL}, computer vision \cite{dualGAN,guo2020closed}, and speech processing \cite{lrspeech}. 
\cite{DL_book} provides a systematic review of research on dual learning. 
Our DLL framework draws inspiration from dual learning and leverages structural duality and probabilistic duality during training and inference. The term `dual learning' may refer to alternative concepts, such as knowledge sharing from multiple parties \cite{MDL} and dual reinforcement of labels \cite{DRL}.

\subsection{Partial Label Learning}
The term ``partial label learning'' is ambiguously used in two contexts. In the first context, it may refer to multi-label classification that deals with partially missing labels in training data \cite{POL,bucak2011multi,missing_label,missing_label2}, where labels often share the same space. In comparison, DLL applies to both classification and regression tasks with irregularly present labels, where dual labels can come from different label spaces. In a different context, the term may also refer to ambiguously labeled learning \cite{chen2020general}, where each training sample is associated with a candidate label set, out of which only one or a few are ground-truth labels \cite{tian2023partial,PML}. 

\subsection{Others}
We compare DLL with a few more works that share similar inspirations. First, DLL relies on a fundamental assumption that the two predicted labels are highly correlated and the correlation can be learnt from data. Similarly, PEMAL \cite{PEMAL} leverages an explicit formula describing the label correlation and directly integrates it into the model architecture. Second, the dual-tower model is designed for direct information exchange between the two towers to complement missing labels. This is similar to X-learner \cite{Xlearner} in causal inference, which cross-trains models by interchanging predicted responses between two treatment groups to complement counterfactuals. Finally, DLL aims at finding solutions to certain equations during alternate inference. Similarly, PINN \cite{PINN} learns a data-driven solution to partial differential equations by enforcing physical laws as regularization terms during training.

\section{Conclusion and Future Work}
We introduce Dual-Label Learning (DLL), a novel framework for training and inference in multi-task learning scenarios where labels are irregularly present. DLL defines a dual-function system that simultaneously fulfills standard supervision, structural duality, and probabilistic duality. At its core, DLL employs a dual-tower model architecture and innovative training and inference algorithms designed to directly model label correlations, maximizing the utility of available labels. Extensive experiments show that DLL consistently outperforms baseline methods.

% Looking ahead, there are opportunities to extend DLL beyond its current limitations. First, DLL is designed on a two-label basis. It can be quickly adapted to the multi-label case via straightforward manipulations such as enumerating label pairs and running DLL for multiple times. However, we foresee that such adaptation could limit the potential of DLL in exploiting label correlations, and we target at a more sophisticated algorithm to tackle the complex label missing patterns arising from an increasing number of labels per sample. Second, the alternative inference scheme raises concern about whether the algorithm converges and converges to the optimal solution, especially when the number of unknowns increases. This points out a research direction toward finding a theoretical guarantee for the convergence of solution in alternative inference, both in the two-label case and the multi-label case. Furthermore, we seek to enhance DLL's capabilities to tackle a wider range of tasks and applications. 

Looking ahead, there are opportunities to extend DLL beyond its current limitations. Currently, DLL is designed for two-label scenarios, and while it can be adapted to multi-label cases by enumerating label pairs and executing multiple times, this approach may limit its ability to fully capture complex label correlations. We aim to develop more sophisticated algorithms to address the intricate patterns that arise with an increasing number of labels. Additionally, the alternate inference scheme used in DLL raises questions about convergence, particularly as the number of unknown labels grows. Future research should focus on establishing theoretical guarantees for convergence in both two-label and multi-label contexts. Furthermore, we seek to enhance DLL's capabilities to address a broader range of tasks and applications.

\section*{Acknowledgments}
This research is supported, in part, by National Natural Science Foundation of China under Grant 62271452 and Hong Kong RGC GRF 16204224.

\appendices 

% \section{List of Notations}
% \input{appendix/notation}

% \section{Training Algorithms} \label{appendix:DLL_train_modes}
% \input{appendix/algorithm}

% \section{Experiment Details} \label{appendix:experiment}
% \input{appendix/experiment}

\bibliographystyle{IEEEtran}
\bibliography{mybibfile}

\newpage

\begin{IEEEbiography}[{\includegraphics[width=1in,height=1.25in,clip,keepaspectratio]{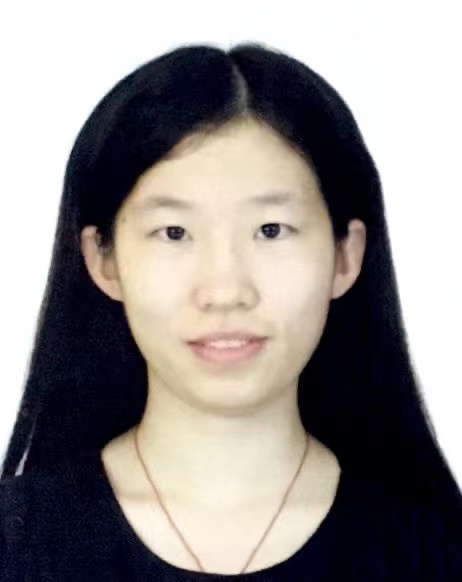}}]{Mingqian Li} currently works as a Post Doc in Zhejiang Lab, Hangzhou, China. She received the BS degree from the National University of Singapore, Singapore in 2016, and the PhD degree from Nanyang Technological University, Singapore in 2023. She has authored or co-authored papers in top tier conferences including AAAI, IJCAI, KDD, ICLR, MobiCom. Her current research interest lies in AI for Science.
\end{IEEEbiography}

\begin{IEEEbiography}[{\includegraphics[width=1in,height=1.25in,clip,keepaspectratio]{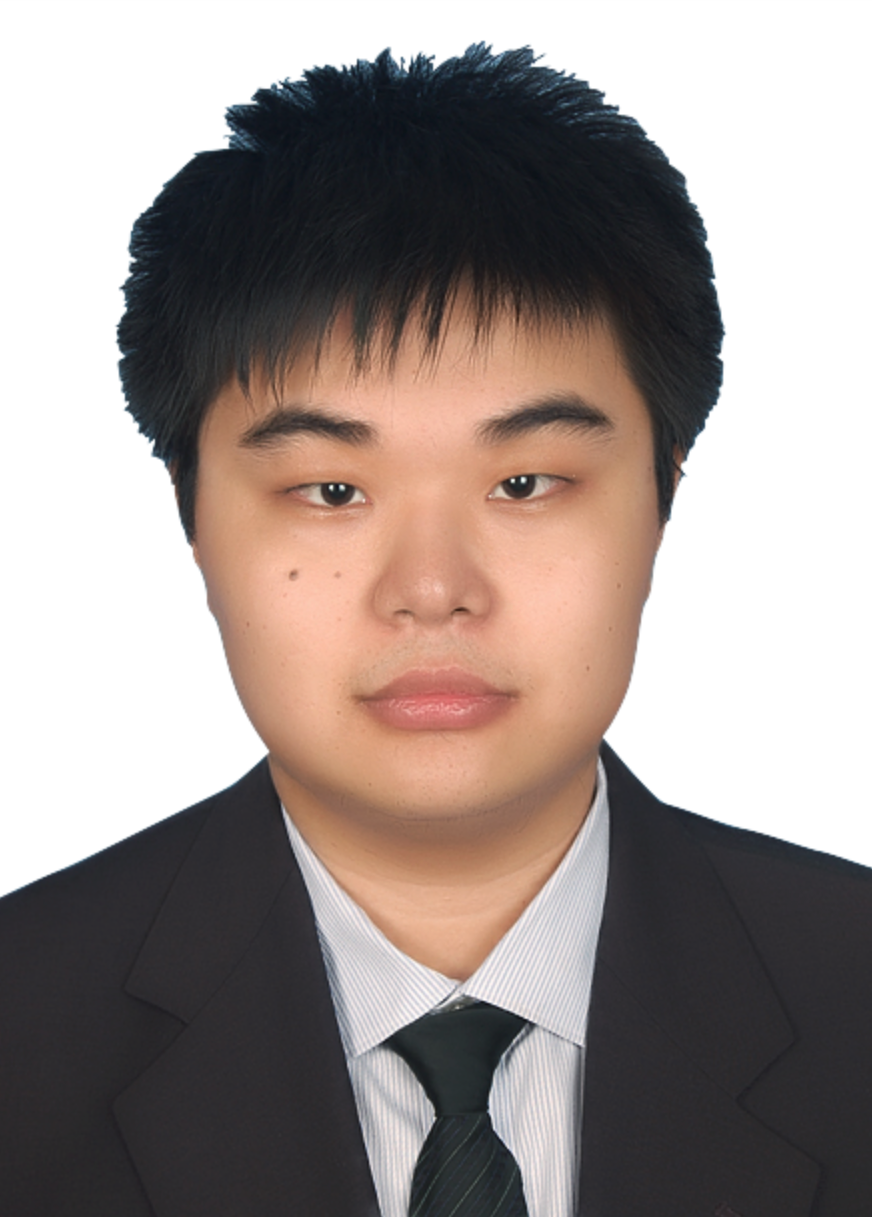}}]{Qiao Han} received the B.S. degree from Tsinghua University, Beijing, China, and the M.S. degree from the National University of Singapore, Singapore. He is currently working as a researcher in Zhejiang Lab, Hangzhou, China. His research interests include Large Language Models and AI for Science. 
\end{IEEEbiography}

\begin{IEEEbiography}[{\includegraphics[width=1in,height=1.25in,clip,keepaspectratio]{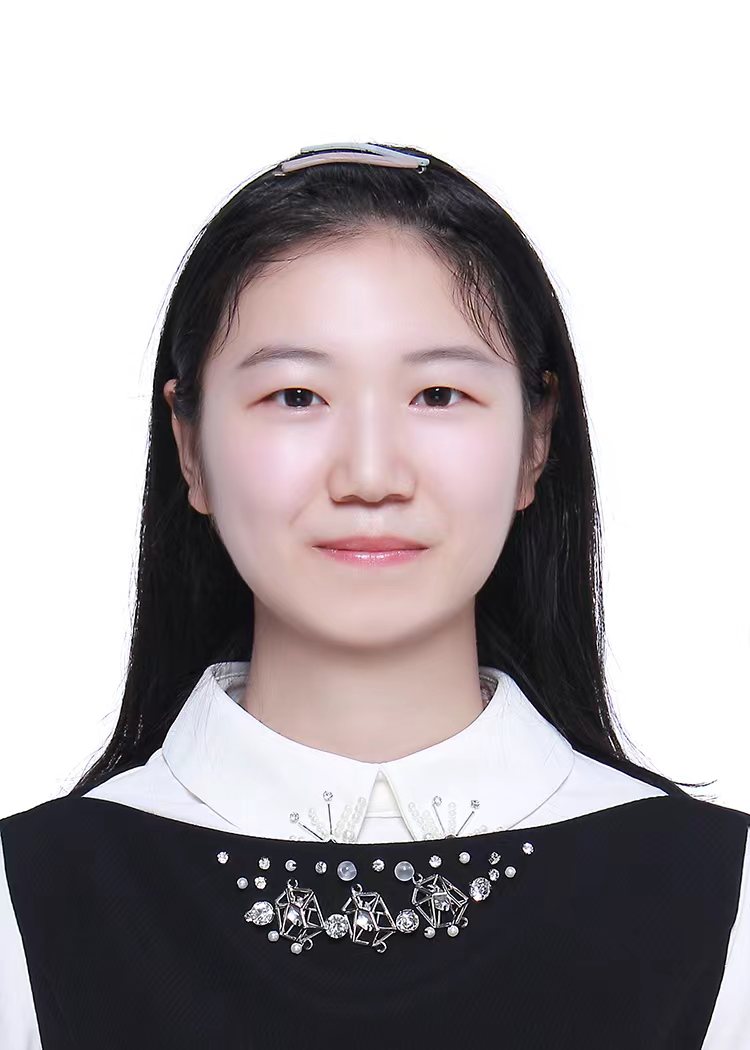}}]{Ruifeng Li} is currently working toward the PhD degree with the College of Computer Science and Technology, Zhejiang University, jointly trained with the Zhejiang Lab. She has authored or co-authored papers in top tier conferences including ICML, KDD, and ICLR. Her interest lies in AI-Driven Drug Design (AIDD) and Large Language Models (LLMs). 
\end{IEEEbiography}

\begin{IEEEbiography}[{\includegraphics[width=1in,height=1.25in,clip,keepaspectratio]{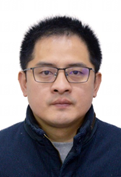}}]{Yao Yang} received the PhD degree from the Department of Physics from Shanghai Jiao Tong University in 2011. He is currently a research expert in Zhejiang Lab, China. His research interests include deep representation learning, federated learning and interdisciplinary topics of machine learning and science. He has published more than 30 research papers in top international journals and conferences.
\end{IEEEbiography}

\begin{IEEEbiography}[{\includegraphics[width=1in,height=1.25in,clip,keepaspectratio]{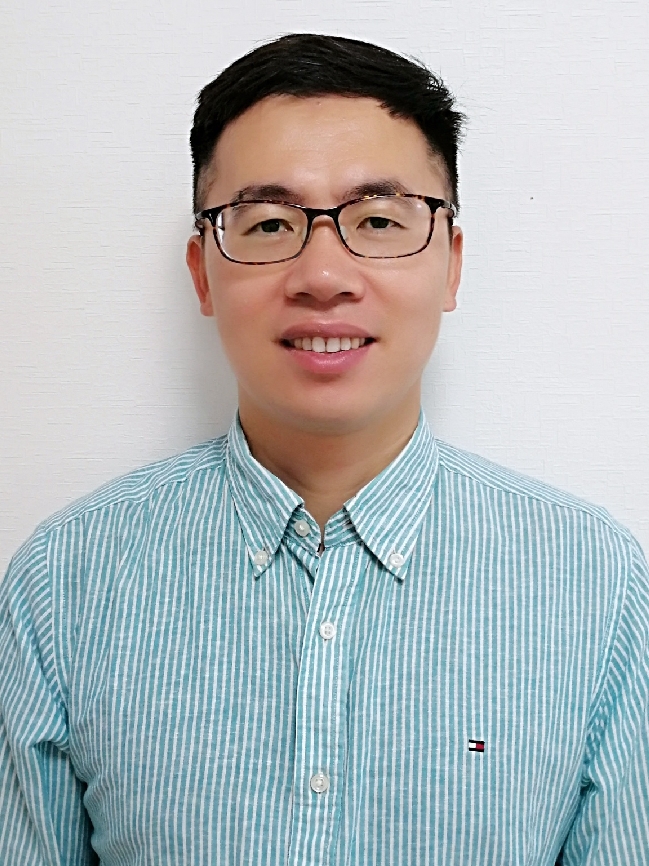}}]{Hongyang Chen}   (M’08-SM’16) received the B.S. and M.S. degrees from Southwest Jiaotong University, Chengdu, China, in 2003 and 2006, respectively, and the Ph.D. degree from The University of Tokyo, Tokyo, Japan, in 2011. From 2011 to 2020, he was a Researcher with Fujitsu Ltd., Tokyo, Japan. He is currently a Senior Research Expert with Zhejiang Lab, China. He has authored or coauthored 130+ refereed journal and conference papers in the IEEE TKDE, TMC, TSP, JSAC, TCOM, SIGMOD, KDD, IJCAI, AAAI, NeurIPS, and has been granted 20+ PCT patents. His research interests include data-driven intelligent systems, graph machine learning, big data mining, and LLM. He was the Editor of the IEEE Journals and a Symposium Chair or Special Session Organizer for some flagship conferences. He was a leading Guest Editor of the IEEE JOURNAL ON SELECTED TOPICS OF SIGNAL PROCESSING on tensor decomposition. Currently, he is an Associate Editor for the IEEE INTERNET OF THINGS JOURNAL. He has been selected as the Distinguished Lecturer of the IEEE Communication Society from 2021 to 2023. He is an adjunct professor with Zhejiang University, China.
\end{IEEEbiography}

\vfill

\end{document}

\title{Dual-Label Learning With Irregularly Present Labels}

\author{Mingqian Li, Qiao Han, Ruifeng Li, Yao Yang,
 Hongyang Chen, ~\IEEEmembership{Senior Member,~IEEE,}
        % <-this % stops a space
\IEEEcompsocitemizethanks{
    % \IEEEcompsocthanksitem This work is supported by XXX (Corresponding author: Hongyang Chen)
    \IEEEcompsocthanksitem  (Corresponding author: Hongyang Chen)
    \IEEEcompsocthanksitem Mingqian Li is with the Research Center for Data Hub and Security, Zhejiang Lab, Hangzhou, China. Email: mingqian.li@zhejianglab.com.
    \IEEEcompsocthanksitem Qiao Han is with the Research Center for Data Hub and Security, Zhejiang Lab, Hangzhou, China. Email: hanq@zhejianglab.com.
    \IEEEcompsocthanksitem Ruifeng Li is with College of Computer Science and Technology, Zhejiang University, Hangzhou, China. Email: lirf@zju.edu.cn.
    \IEEEcompsocthanksitem Yao Yang is with the Research Center for Data Hub and Security, Zhejiang Lab, Hangzhou, China. Email: yangyao@zhejianglab.com.
    \IEEEcompsocthanksitem Hongyang Chen is with the Research Center for Data Hub and Security, Zhejiang Lab, Hangzhou, China. Email: dr.h.chen@ieee.org.}
% <-this % stops a space
\thanks{Manuscript received April 19, 2021; revised August 16, 2021.}
}

\markboth{Journal of \LaTeX\ Class Files,~Vol.~14, No.~8, August~2021}%
{Shell \MakeLowercase{\textit{et al.}}: A Sample Article Using IEEEtran.cls for IEEE Journals}


% \maketitle

\appendices 

\section{List of Notations}
Table \ref{tab:notation} summarizes the notations adopted in this paper.

\begin{table}[ht] 
    \centering
     \begin{tabular}{c|c}
         Symbol & Meaning  \\
         \hline
         $\Theta_{f}, \Theta_{g}$ & parameter spaces of $f$ and $g$\\
         $\alpha$ & coefficients\\
         $\delta$ & threshold\\
         $\theta_0$ & shared parameters of $f$ and $g$\\
         $\theta_1$ & parameters of Tower 1 and Embedding 1\\
         $\theta_2$ & parameters of Tower 2 and Embedding 2\\
         $\theta_{f}$ & parameters of $f$ \\
         $\theta_{g}$ & parameters of $g$ \\
         $\lambda_{11}, \lambda_{22}$ & reconstruction loss coefficients \\
         $\lambda_{12}, \lambda_{21}$ & supervision loss coefficients \\
         $\lambda_{d}$ & duality loss coefficients \\
         $\sigma$ & Rademacher variables\\
         
         $\mathcal{D}$ & probability distribution \\
         $\mathcal{F}, \mathcal{G}$ & primal function space, dual function space\\
         $\mathcal{H}$ & sub-space of the problem space\\
         $\mathcal{S}$ & dataset \\
         $\mathcal{S}_l$ & full-labeled sub-dataset\\
         $\mathcal{S}_u$ & unlabeled sub-dataset\\
         $\mathcal{S}_1$ & semi-labeled sub-dataset with $y_1$ labels\\
         $\mathcal{S}_2$ & semi-labeled sub-dataset with $y_2$ labels\\
         $\mathcal{X}, \mathcal{Y}_1, \mathcal{Y}_2$ & feature space, $y_1$ label space, $y_2$ label space \\

         $B$ & batch size \\
         $G_0, G_{1}, G_{2}$ & gradients of the parameters \\
         $I_l, I_1, I_2, I_u$ & indices for the sub-datasets\\
         $L$ & number of iterations \\
         $M$ & the pre-trained multi-task model \\
         $N$ & number of epochs \\
         N.A. & the label is missing \\
         $Opt$ & optimizer \\
         $P_f, P_g, P_M$ & probabilistic outputs of $f$, $g$, and $M$ \\
         $R$ & expected risk\\
         $R_n$ & empirical risk\\
         $R_n^{DLL}$ & Rademacher complexity of DLL\\
         $T_1, T_2$ & tasks \\
         $U$ & label presence vector space \\

         $d$ & duality loss \\
         $f, g$ & primal function, dual function \\
         $i$ & sample index \\
         $k$ & complexity of a feed-forward pass \\
         $l_1, l_2, l_{f, g}, l_d$ & loss functions \\
         $n$ & total number of samples in $\mathcal{S}$ \\
         $r_1$ & reconstruction loss on $y_1$ \\
         $r_2$ & reconstruction loss on $y_2$ \\
         $s_1$ & supervision loss on $y_1$ \\
         $s_2$ & supervision loss on $y_2$ \\
         $u$ & vector indicator of present labels \\
         $x$ & features \\
         $y_0$ & label initialization \\
         $y_1, y_2$ & labels \\
         $\hat{}$ & estimated value \\
         $\nabla$ & gradient operator \\
         
    \end{tabular}
    \caption{Notations.}
    \label{tab:notation}
\end{table}

\section{Training Algorithms} \label{appendix:DLL_train_modes}
We detail the algorithm of each training mode (a-c, f) as if the dual-tower model is trained on this mode separately.

\begin{algorithm}[t]
\caption{Algorithms of Different Training Modes} \label{algo:DLL_train_modes}

\begin{flushleft}
\textbf{Input:} 
Training datasets $\mathcal{S}_l, \mathcal{S}_1, \mathcal{S}_2$; supervision loss coefficients $\lambda_{12}, \lambda_{21}$; reconstruction loss coefficients $\lambda_{11}, \lambda_{22}$; duality loss coefficients $\lambda_{d}$; optimizer $Opt$ 

\textbf{Output:} 
the dual-tower model parameters $\hat{\theta}_{0}, \hat{\theta}_{1}, \hat{\theta}_{2}$
\end{flushleft}

\begin{algorithmic}[1]

\STATE \textbf{if} (f) multi-task \textbf{then} 
\STATE \hspace{0.5cm} Pre-train the multi-task model $M$ on $\mathcal{S}_l$: 
\STATE \hspace{0.5cm} $M \sim P(y_1|x), P(y_2|x)$;
\STATE \hspace{0.5cm} \textbf{return} $M$

\STATE \textbf{elif} (a) supervised learning \textbf{then}
\STATE \hspace{0.5cm} Sample $(x^{(i)}, y_1^{(i)}, y_2^{(i)})$ from $\mathcal{S}_l$;
% \STATE \hspace{0.5cm} Calculate the gradients:
\STATE \hspace{0.5cm} $G_0 = \nabla_{\theta_{0}} [ 
\lambda_{12} l_2 (f(x^{(i)}, y_1^{(i)}; \hat{\theta}_{2}, \hat{\theta}_{0}), y_2^{(i)})$
\STATE \hspace{1cm} $+ \lambda_{21} l_1 (g(x^{(i)}, y_2^{(i)}; \hat{\theta}_{1}, \hat{\theta}_{0}), y_1^{(i)}) ]$;
\STATE \hspace{0.5cm} $G_{1} = \nabla_{\theta_{1}} 
\lambda_{21} l_1 (g(x^{(i)}, y_2^{(i)}; \hat{\theta}_{1}, \hat{\theta}_{0}), y_1^{(i)})$;
\STATE \hspace{0.5cm} $G_{2} = \nabla_{\theta_{2}} 
\lambda_{12} l_2 (f(x^{(i)}, y_1^{(i)}; \hat{\theta}_{2}, \hat{\theta}_{0}), y_2^{(i)})$;
% \STATE \hspace{0.5cm} Update the parameters:
\STATE \hspace{0.5cm} $\hat{\theta}_{0} \leftarrow Opt(\hat{\theta}_{0}, G_0)$;
\STATE \hspace{0.5cm} $\hat{\theta}_{1} \leftarrow Opt(\hat{\theta}_{1}, G_{1})$;
\STATE \hspace{0.5cm} $\hat{\theta}_{2} \leftarrow Opt(\hat{\theta}_{2}, G_{2})$;
% \STATE \hspace{0.5cm} $\hat{\theta}_{0}, \hat{\theta}_{1}, \hat{\theta}_{2} \leftarrow Opt(\hat{\theta}_{0}, G_0), Opt(\hat{\theta}_{1}, G_{1}), Opt(\hat{\theta}_{2}, G_{2})$;
\STATE \hspace{0.5cm} \textbf{return} $\hat{\theta}_{0}, \hat{\theta}_{1}, \hat{\theta}_{2}$

\STATE \textbf{elif} (b1) learning from structural duality on $\mathcal{S}_1$ \textbf{then}
\STATE \hspace{0.5cm} Sample $(x^{(i)}, y_1^{(i)})$ from $\mathcal{S}_1$;
\STATE \hspace{0.5cm} Estimate $\hat{y}_2^{(i)} = f(x^{(i)}, y_1^{(i)}; \hat{\theta}_{2}, \hat{\theta}_{0})$
% \STATE \hspace{0.5cm} Calculate the gradients:
\STATE \hspace{0.5cm} $G_{0} = \nabla_{\theta_{0}} 
\lambda_{11} l_1 (g(x^{(i)}, \hat{y}_2^{(i)}; \hat{\theta}_{1}, \hat{\theta}_{0}), y_1^{(i)})$;
\STATE \hspace{0.5cm} $G_{1} = \nabla_{\theta_{1}} 
\lambda_{11} l_1 (g(x^{(i)}, \hat{y}_2^{(i)}; \hat{\theta}_{1}, \hat{\theta}_{0}), y_1^{(i)})$;
% \STATE \hspace{0.5cm} Update the parameters:
\STATE \hspace{0.5cm} $\hat{\theta}_{0}, \hat{\theta}_{1} \leftarrow Opt(\hat{\theta}_{0}, G_0), Opt(\hat{\theta}_{1}, G_{1})$;
\STATE \hspace{0.5cm} \textbf{return} $\hat{\theta}_{0}, \hat{\theta}_{1}$

\STATE \textbf{elif} (b2) learning from structural duality on $\mathcal{S}_2$ \textbf{then}
\STATE \hspace{0.5cm} Repeat (b1) on $\mathcal{S}_2$ symmetrically.

\STATE \textbf{elif} (c) learning from probabilistic duality \textbf{then}
% \STATE \hspace{0.5cm} \textbf{repeat}
\STATE \hspace{0.5cm} $M \gets$ (f) multi-task;
\STATE \hspace{0.5cm} Sample $(x^{(i)}, y_1^{(i)}, y_2^{(i)})$ from $\mathcal{S}_l \cup \mathcal{S}_1 \cup \mathcal{S}_2$;
\STATE \hspace{0.5cm} \textbf{if} $y_1^{(i)}$ is N.A. \textbf{then} estimate $y_1^{(i)}$ with $g$;
\STATE \hspace{0.5cm} \textbf{if} $y_2^{(i)}$ is N.A. \textbf{then} estimate $y_2^{(i)}$ with $f$;
\STATE \hspace{0.5cm} Estimate $\hat{p}_1 := P_M(y_1^{(i)}|x^{(i)}), \hat{p}_2 := P_M(y_2^{(i)}|x^{(i)})$;
% \STATE \hspace{0.5cm} Calculate the gradients:
\STATE \hspace{0.5cm} $loss = (\log \hat{p}_1 - \log \hat{p}_2$
\STATE \hspace{1.5cm} $ + \log P_f(y_2^{(i)}|x^{(i)}, y_1^{(i)};\hat{\theta}_{2}, \hat{\theta}_0)$
\STATE \hspace{1.5cm} $ - \log P_g(y_1^{(i)}|x^{(i)}, y_2^{(i)}; \hat{\theta}_{1}, \hat{\theta}_0))$
\STATE \hspace{0.5cm} $G_0, G_{1}, G_{2} =  \nabla_{\theta_{0}, \theta_{1}, \theta_{2}} (\lambda_{d} \cdot loss)$
% \STATE \hspace{0.5cm} Update the parameters:
\STATE \hspace{0.5cm} $\hat{\theta}_{0} \leftarrow Opt(\hat{\theta}_{0}, G_0)$;
\STATE \hspace{0.5cm} $\hat{\theta}_{1} \leftarrow Opt(\hat{\theta}_{1}, G_{1})$;
\STATE \hspace{0.5cm} $\hat{\theta}_{2} \leftarrow Opt(\hat{\theta}_{2}, G_{2})$;
\STATE \hspace{0.5cm} \textbf{return} $\hat{\theta}_{0}, \hat{\theta}_{1}, \hat{\theta}_{2}$

\end{algorithmic}
\end{algorithm}

\section{Experiment Details} \label{appendix:experiment}

\subsection{Implementation details}

\subsubsection{Data preprocessing}
In the Tox21 dataset, the RDKit topological fingerprints are extracted from SMILES with the open-source Python RDKit library\footnote{https://www.rdkit.org/} and used as features. The train-val-test split is 64:16:20 across all the three datasets for evaluation purposes. In real-world scenarios, single-label prediction should apply to sub-datasets $\mathcal{S}_1$ and $\mathcal{S}_2$, and double-label prediction should apply to sub-dataset $\mathcal{S}_u$; while for experiments, we reserve 20\% of the samples in $\mathcal{S}_l$ for evaluation purpose, by properly masking the labels and evaluating the model with the ground truth labels. In either task, the prediction is on both labels $y_1$ and $y_2$.

\subsubsection{Models}
The DLL framework offers flexibility in the choice of model architecture. In our experiments, all modules within the dual-tower model, including the towers themselves, are implemented using multilayer perceptrons (MLPs). The model is built in PyTorch and is trained on a Single GPU (NVIDIA GeForce RTX 2080 Ti) using the stochastic gradient descent (SGD) optimizer (for $Opt$). For baseline approaches, model selection is implemented with AutoGluon \cite{agtabular}. 

\subsubsection{Training}
Loss functions are configured during training. In binary classification tasks, cross entropy loss is applied to $l_1$ and $l_2$, and MSE loss is applied to $l_{d}$. In regression tasks, MSE loss is used to configure all loss functions. 
The total loss function is the weighted summed with heuristically decided loss coefficients 
$(\lambda_{11}, \lambda_{21}, \lambda_{12}, \lambda_{22}, \lambda_{d} ) = (2, 2, 1, 1, 0.2)$ in Tox21, 
$(\lambda_{11}, \lambda_{21}, \lambda_{12}, \lambda_{22}, \lambda_{d} ) = (1, 1, 1, 1, 0)$ in HIGGS, 
and $(\lambda_{11}, \lambda_{21}, \lambda_{12}, \lambda_{22}, \lambda_{d} ) = (2, 2, 1, 1, 0)$ in MOF. Moreover, positive and negative samples in Tox21 are re-weighted by 0.7:0.3 in calculating losses to mitigate the class imbalance issue. The maximum number of training epochs is set as $N = 100$ in all three datasets. The batch size is set as $B = 4$ in Tox21 and HIGGS, and $B = 1$ in MOF.

\subsubsection{Inference}
In alternate inference, labels are initialized as $y_0 = 0.5$ in binary classification, and $y_0 = 1.0$ in regression, according to our observation of the dataset distribution. We conduct alternate inference for $L = 1,000$ iterations without exerting any convergence threshold on earlier stopping. In fact, setting such a threshold could have further reduced the computational cost.

\subsection{Source code}
The source code and experiment data, upon paper acceptance, will be made publicly available on GitHub.

\subsection{Experiment results} \label{appendix:exp_result}

Table \ref{tab:baseline_tox21}, Table \ref{tab:baseline_higgs} and Table \ref{tab:baseline_mof} provide full results of baseline comparison on the three datasets, including all evaluation metrics. 

\begin{table*}[htb]
\setlength{\tabcolsep}{1.2mm}
\centering
% \small
  \begin{tabular}{cc|l|cccc|cccc}
    \multicolumn{9}{l}{\textbf{Tox21:}} \\
    \midrule
     & & & \multicolumn{4}{c|}{$y_1$} & \multicolumn{4}{c}{$y_2$}\\ 
    \cmidrule{4-11} &&\textbf{Method}&Accuracy$\uparrow$&Precision$\uparrow$&Recall$\uparrow$&F1-score$\uparrow$&Accuracy$\uparrow$&Precision$\uparrow$&Recall$\uparrow$&F1-score$\uparrow$\\ 
    \midrule
    \midrule
    \multirow{7}{*}{\rotatebox{90}{\textbf{Single-label}}} &\multirow{7}{*}{\rotatebox{90}{\textbf{prediction}}}
    &ID   &0.79$\pm$0.02	&0.40$\pm$0.04	&0.44$\pm$0.10	&0.41$\pm$0.05	&0.84$\pm$0.02	&0.51$\pm$0.08	&0.55$\pm$0.11	&0.52$\pm$0.04\\
    &&COL  &0.78$\pm$0.04	&0.33$\pm$0.06	&0.46$\pm$0.09	&0.38$\pm$0.02	&0.87$\pm$0.01	&0.47$\pm$0.05	&0.53$\pm$0.05	&0.50$\pm$0.04\\
    &&SSL  &0.83$\pm$0.01	&0.53$\pm$0.04	&0.33$\pm$0.05	&0.40$\pm$0.04	&0.87$\pm$0.01	&0.64$\pm$0.06	&0.48$\pm$0.06	&0.54$\pm$0.03\\
    &&LS   &0.79$\pm$0.02	&0.40$\pm$0.04	&0.44$\pm$0.10	&0.41$\pm$0.05	&\textbf{0.91$\pm$0.01}	&0.66$\pm$0.05	&0.53$\pm$0.09	&0.59$\pm$0.05\\
    &&DSML       &\textbf{0.87$\pm$0.01}	&\textbf{0.74$\pm$0.09}	&0.27$\pm$0.03	&0.40$\pm$0.03	&0.90$\pm$0.01	&\textbf{0.69$\pm$0.07}	&0.53$\pm$0.06	&0.60$\pm$0.05\\
   &&DSML (rev) &0.84$\pm$0.03	&0.44$\pm$0.07	&\textbf{0.50$\pm$0.05}	&0.46$\pm$0.04	&0.91$\pm$0.01	&0.64$\pm$0.06	&0.48$\pm$0.07	&0.54$\pm$0.05\\
    \cmidrule{3-11}
    &&\textbf{DLL (ours)}	&0.85$\pm$0.01	&0.58$\pm$0.07	&0.40$\pm$0.05	&\textbf{0.47$\pm$0.04}	&0.90$\pm$0.01	&0.68$\pm$0.04	&\textbf{0.57$\pm$0.01}	&\textbf{0.62$\pm$0.01}\\
    \midrule
    % \multicolumn{9}{l} {} \\
    % \multicolumn{9}{l}{\textbf{Tox21: double-label prediction}} \\
    % \midrule
    % { } & \multicolumn{4}{c|}{$y_1$} & \multicolumn{4}{c}{$y_2$}\\ 
    % \cmidrule{2-9} \textbf{Method}&Accuracy$\uparrow$&Precision$\uparrow$&Recall$\uparrow$&F1-score$\uparrow$&Accuracy$\uparrow$&Precision$\uparrow$&Recall$\uparrow$&F1-score$\uparrow$\\ 
    \midrule
    \multirow{7}{*}{\rotatebox{90}{\textbf{Double-label}}} &\multirow{7}{*}{\rotatebox{90}{\textbf{prediction}}}
    &ID   &0.79$\pm$0.02	&0.40$\pm$0.04	&0.44$\pm$0.10	&0.41$\pm$0.05	&0.84$\pm$0.02	&0.51$\pm$0.08	&0.55$\pm$0.11	&0.52$\pm$0.04\\
    &&COL  &0.78$\pm$0.04	&0.33$\pm$0.06	&0.46$\pm$0.09	&0.38$\pm$0.02	&0.87$\pm$0.01	&0.47$\pm$0.05	&0.53$\pm$0.05	&0.50$\pm$0.04\\
    &&SSL   &0.83$\pm$0.01	&0.53$\pm$0.04	&0.33$\pm$0.05	&0.40$\pm$0.04	&0.87$\pm$0.01	&0.64$\pm$0.06	&0.48$\pm$0.06	&0.54$\pm$0.03\\
    &&LS   &0.79$\pm$0.02	&0.40$\pm$0.04	&0.44$\pm$0.10	&0.41$\pm$0.05	&0.90$\pm$0.01	&0.69$\pm$0.06	&0.37$\pm$0.06	&0.48$\pm$0.04\\
    &&DSML    &\textbf{0.87$\pm$0.01}	&\textbf{0.64$\pm$0.06}	&0.28$\pm$0.04	&0.39$\pm$0.03	&0.89$\pm$0.01	&0.54$\pm$0.05	&0.61$\pm$0.08	&0.57$\pm$0.03\\
   &&DSML (rev) &0.85$\pm$0.01	&0.53$\pm$0.06	&0.37$\pm$0.04	&0.44$\pm$0.03	&\textbf{0.91$\pm$0.01}	&\textbf{0.76$\pm$0.06}	&0.45$\pm$0.05	&0.57$\pm$0.04\\
    \cmidrule{3-11}
    &&\textbf{DLL (ours)}	&0.82$\pm$0.01	&0.47$\pm$0.04	&\textbf{0.47$\pm$0.03}	&\textbf{0.47$\pm$0.03}	&0.87$\pm$0.01	&0.54$\pm$0.04	&\textbf{0.61$\pm$0.04}	&\textbf{0.57$\pm$0.03} \\
    \midrule
  \end{tabular}
  \caption{Full results of baseline comparison on the Tox21 dataset. Best average results are bolded.}
  \label{tab:baseline_tox21}
\end{table*}

\begin{table*}[htb]
\setlength{\tabcolsep}{1.2mm}
\centering
% \small
  \begin{tabular}{l|c|c|c|c}
  \multicolumn{1}{l}{\textbf{HIGGS:}}\\
    \midrule
    \multicolumn{1}{l|}{}& \multicolumn{2}{c|}{\textbf{Single-label prediction}}& \multicolumn{2}{c}{\textbf{Double-label prediction}}\\
    \cmidrule{2-5}
    \textbf{Method}& \multicolumn{1}{c|}{$y_1$} & \multicolumn{1}{c|}{$y_2$} & \multicolumn{1}{c|}{$y_1$} & \multicolumn{1}{c}{$y_2$}\\ 
    % \cmidrule{2-13} \textbf{Missing rate}&10\%&20\%&30\%&40\%&50\%&60\%&10\%&20\%&30\%&40\%&50\%&60\%\\ 
    \midrule
    ID  &31.1\%$\pm$1.6\%	&18.9\%$\pm$0.5\%	&31.1\%$\pm$1.6\%	&18.9\%$\pm$0.5\%\\
    COL &31.7\%$\pm$1.5\%	&19.2\%$\pm$0.4\%	&31.7\%$\pm$1.5\%	&19.2\%$\pm$0.4\%\\
    SSL &30.2\%$\pm$1.4\%	&18.6\%$\pm$0.4\%	&30.2\%$\pm$1.4\%	&18.6\%$\pm$0.4\%\\
    LS &31.1\%$\pm$1.6\%	&17.1\%$\pm$0.7\%	&31.1\%$\pm$1.6\%	&19.0\%$\pm$0.5\%\\
    DSML &29.8\%$\pm$1.1\%	&17.3\%$\pm$1.2\%	&29.9\%$\pm$1.0\%	&18.5\%$\pm$0.6\%\\
    DSML (rev) &28.3\%$\pm$1.1\%	&17.7\%$\pm$0.5\%	&30.2\%$\pm$1.4\%	&18.8\%$\pm$0.6\%\\
    \cmidrule{1-5}
    \textbf{DLL (ours)} &\textbf{25.0\%$\pm$1.3\%}	&\textbf{17.0\%$\pm$0.1\%}	&\textbf{25.9\%$\pm$1.9\%}	&\textbf{17.2\%$\pm$0.3\%} \\
    \cmidrule{1-5} 

  \end{tabular}
  \caption{Full results of baseline comparison on the HIGGS dataset. The evaluation metric is MAPE ($\downarrow$). Best average results are bolded.}
  \label{tab:baseline_higgs}
\end{table*}

\begin{table*}[htb]
\setlength{\tabcolsep}{1.2mm}
\centering
% \small
  \begin{tabular}{l|c|c|c|c}
  \multicolumn{1}{l}{\textbf{MOF:}}\\
    \midrule
    \multicolumn{1}{l|}{}& \multicolumn{2}{c|}{\textbf{Single-label prediction}}& \multicolumn{2}{c}{\textbf{Double-label prediction}}\\
    \cmidrule{2-5}
    \textbf{Method}& \multicolumn{1}{c|}{$y_1$} & \multicolumn{1}{c|}{$y_2$} & \multicolumn{1}{c|}{$y_1$} & \multicolumn{1}{c}{$y_2$}\\ 
    % \cmidrule{2-13} \textbf{Missing rate}&10\%&20\%&30\%&40\%&50\%&60\%&10\%&20\%&30\%&40\%&50\%&60\%\\ 
    \midrule
    ID  &24.0\%$\pm$0.9\%	&46.0\%$\pm$2.7\%	&24.0\%$\pm$0.9\%	&46.0\%$\pm$2.7\%\\
    COL &23.9\%$\pm$0.7\%	&47.2\%$\pm$4.4\%	&23.9\%$\pm$0.7\%	&47.2\%$\pm$4.4\%\\
    SSL &23.4\%$\pm$1.2\%	&45.6\%$\pm$2.0\%	&23.4\%$\pm$1.2\%	&45.6\%$\pm$2.0\%\\
    LS &24.0\%$\pm$0.9\%	&46.8\%$\pm$2.6\%	&24.0\%$\pm$0.9\%	&48.2\%$\pm$4.3\%\\
    DSML &23.8\%$\pm$1.7\%	&43.5\%$\pm$2.0\%	&24.2\%$\pm$1.1\%	&44.8\%$\pm$2.0\%\\
    DSML (rev) &23.3\%$\pm$0.6\%	&45.3\%$\pm$2.0\%	&23.7\%$\pm$0.8\%	&45.8\%$\pm$2.3\%\\
    \cmidrule{1-5}
    \textbf{DLL (ours)} &\textbf{22.3\%$\pm$0.6\%}	&\textbf{42.7\%$\pm$1.4\%}	&\textbf{22.8\%$\pm$0.5\%}	&\textbf{43.8\%$\pm$1.1\%} \\
    \cmidrule{1-5} 

  \end{tabular}
  \caption{Full results of baseline comparison on the MOF dataset. The evaluation metric is MAPE ($\downarrow$). Best average results are bolded.}
  \label{tab:baseline_mof}
\end{table*}

\bibliographystyle{IEEEtran}
\bibliography{mybibfile}

\vfill